\pdfoutput=1
\documentclass[10pt,twocolumn,letterpaper]{article}
\usepackage[pagenumbers]{cvpr}
\usepackage{amsmath,amssymb,amsthm,booktabs,graphicx,xcolor,colortbl,multirow}
\usepackage{tikz}\usetikzlibrary{arrows.meta,calc,positioning,shapes.geometric,decorations.pathreplacing}
\usepackage[pagebackref,breaklinks,colorlinks]{hyperref}
\definecolor{cvprblue}{rgb}{0.21,0.49,0.74}\hypersetup{allcolors=cvprblue}
\usepackage{enumitem}
\usepackage[capitalize]{cleveref}

\usepackage[numbers,sort&compress]{natbib}
\definecolor{ourspastel}{RGB}{247,236,230}\definecolor{goodgreen}{RGB}{220,242,222}\definecolor{tieyellow}{RGB}{255,246,200}\definecolor{poorred}{RGB}{252,222,222}
\definecolor{ours}{RGB}{169,74,34}
\definecolor{ink}{RGB}{58,58,56}
\definecolor{sblue}{RGB}{34,98,175}
\usepackage[most]{tcolorbox}
\theoremstyle{plain}\newtheorem{theorem}{Theorem}\newtheorem{proposition}{Proposition}
\newtheorem{lemmaraw}{Lemma}
\definecolor{thmtint}{RGB}{252,246,241}
\tcbset{arborthm/.style={enhanced, colback=thmtint, colframe=thmtint, boxrule=0pt, sharp corners, left=5pt, right=5pt, top=2pt, bottom=2pt, before skip=5pt, after skip=5pt}}
\usepackage{needspace}\usepackage{mdframed}\mdfsetup{hidealllines=true,nobreak=true,backgroundcolor=thmtint,innerleftmargin=5pt,innerrightmargin=5pt,innertopmargin=3pt,innerbottommargin=3pt,skipabove=5pt,skipbelow=5pt,splittopskip=0pt,splitbottomskip=0pt}
\newenvironment{lemma}[1][]{\begin{lemmaraw}[#1]}{\end{lemmaraw}}
\crefname{lemmaraw}{Lemma}{Lemmas}\Crefname{lemmaraw}{Lemma}{Lemmas}
\newcommand{\R}{\mathbb{R}}

\def\confName{CVPR}\def\confYear{2027}
\makeatletter\AtBeginDocument{\@ifundefined{bibsep}{}{\setlength{\bibsep}{0pt plus 0.5pt}}}\makeatother
\begin{document}
\raggedbottom

\title{Arbor: When Labels Are Not Coarsened at Random}

\author{Faisal Mohammad$^{1,2}$\qquad Mohamed Younis$^{1}$\\
$^{1}$University of Maryland, Baltimore County\\
$^{2}$Johns Hopkins University Applied Physics Laboratory\\
{\tt\small \{faisalm2,\,younis\}@umbc.edu}}
\maketitle
\begin{abstract}
Merging datasets labelled at different levels of a taxonomy leaves some images with only a coarse label, and some classes with no fine-grained label at all. Standard training scores a coarse label by the total probability of the classes under it. Coarse labels name superclasses, and the standard objective is exact only when every class within a superclass is equally likely to receive a coarse label. We call a class \emph{novel} if it is never labelled finely, and \emph{known} otherwise. A novel class violates the assumption, and the classifier assigns its images to a known sibling. We propose Arbor, which models how labels become coarse. In each superclass with a novel class, Arbor adds a virtual novel class, and a logistic gate divides the superclass's probability between it and the known classes. The gate is trained with the likelihood of the observed labels under this model, the coarsening likelihood. We prove that at its population optimum the gate equals the Elkan--Noto correction of positive-unlabelled learning, and show that training the correction into the classifier improves on applying it after training. Across Cars, CUB and Aircraft with 90\% coarse labels, Arbor correctly assigns 76--85\% of novel-class images to their superclass, more than hierarchical classifiers, hierarchical novelty detection and the Elkan--Noto correction. Its top-1 accuracy over the leaf classes on known-class images is within 1.6 points of the best baseline, although its final predictions return the superclass more often. On the more challenging iNaturalist-2019 benchmark, Arbor improves novel-class accuracy from 46\% (uniform target, the strongest baseline) to 68\%.
\end{abstract}
\begin{figure}[t!]
\centering
\includegraphics[width=0.68\linewidth]{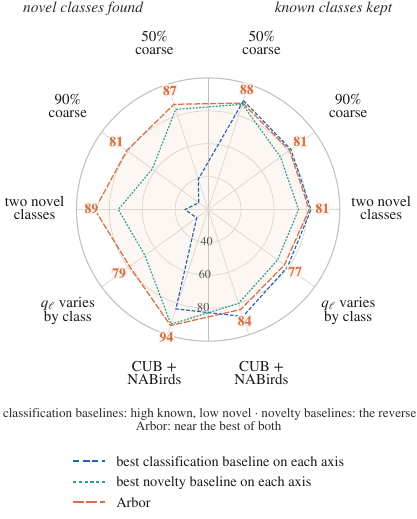}
\caption{\textbf{Classification baselines keep known classes but find few novel ones, and novelty baselines do the reverse. Arbor is close to the better of the two on every axis.} The radial axis starts at $20\%$. Right half: known-class leaf accuracy, the top-1 accuracy over leaves, which does not depend on the operating point. Left half: novel-class accuracy, the share of novel-class images assigned to their superclass. Each protocol axis (\cref{sec:setup}) averages Cars, CUB, CIFAR-100 and Aircraft. The bottom pair is the real CUB--NABirds merge (\cref{tab:merge}).}
\label{fig:teaser}
\end{figure}
\section{Introduction}
In a taxonomy of ships, container ships and bulk carriers are two types of cargo ship. Suppose one dataset labels its images ``container ship'' and a second dataset labels the same kind of image only ``cargo ship''. After merging, some images carry a type label and others only the label ``cargo ship''. Bulk carriers appear only in the second dataset: every bulk carrier is labelled ``cargo ship'', and no image is labelled ``bulk carrier''. Standard training scores a coarse label by the total probability of the classes under it~\cite{cour2011partial,valmadre2022hierarchical}, so the loss on a bulk carrier image is the same however its probability is split among the cargo-ship types. Nothing rewards leaving the type open, and only the known types ever receive labels, so they take the probability. The classifier calls a bulk carrier a container ship when ``cargo ship, unknown type'' would be correct.

The standard objective is the exact likelihood only if the labels are \emph{coarsened at random}~\cite{heitjan1991ignorability,gill1997coarsening}: every fine-grained class under a superclass such as ``cargo ship'' must be equally likely to have been labelled with the superclass instead. We call a class \emph{known} if it receives fine-grained labels at least some of the time, and \emph{novel} if its images only ever carry the superclass label~\cite{lee2018hierarchical}. A novel class breaks the condition: it is always labelled with the superclass, its known siblings only sometimes. Container ships are known and bulk carriers are novel, and both are listed in the taxonomy. Given the coarsening probabilities, the labels determine the probability of each known class, but only the total probability of the novel classes~\cite{liu2014learnability}.

\begin{figure*}[t]
\centering
\includegraphics[width=\textwidth]{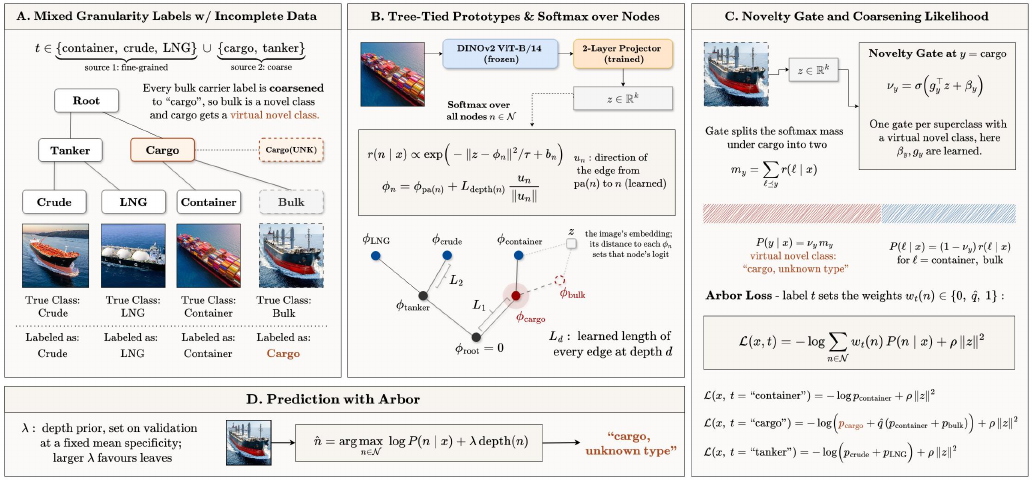}
\
\label{fig:overview}
\end{figure*}

Our method, Arbor, learns that total. In each superclass that contains a novel class, Arbor adds a \emph{virtual novel class}, one extra class standing for any type under the superclass that no label names. A logistic gate trained with the coarsening likelihood of \cref{sec:rule} divides the superclass's probability between the virtual novel class and the known classes (\cref{fig:overview}). We prove that at its optimum the gate equals the Elkan--Noto correction~\cite{elkan2008pu}, the standard positive-unlabelled estimate (\cref{thm:pooled}). Training the correction into the classifier improves on applying it after training (\cref{sec:results}).

Existing methods are good at one of the two tasks, classifying the known classes and recognising the novel ones, and poor at the other (\cref{fig:teaser}). Hierarchical classifiers~\cite{deng2012hedging,goren2024hsc,valmadre2022hierarchical,zeng2024hypstructure,barz2019hierarchy,wu2020deeprtc,chu2026hieruq,park2026freegrain} learn the known classes but miss most novel classes (\cref{fig:examples}). Hierarchical novelty detection~\cite{lee2018hierarchical} finds novel classes but loses known-class accuracy (\cref{sec:related}). Arbor differs from it in where the virtual novel class lives: its classifier is a softmax over the nodes of the taxonomy~\cite{valmadre2022hierarchical}, and its gate is a separate logistic unit on the same embedding (\cref{sec:ablation}).
\paragraph{Contributions.}
\begin{itemize}[leftmargin=1.2em,itemsep=2pt,topsep=2pt]
\item We show that a class without fine-grained labels violates coarsening at random and turns each affected superclass into a positive-unlabelled problem (\cref{sec:theory}).
\item We propose Arbor and prove that the optimum of its gate is the Elkan--Noto correction and that the labels bound the coarsening probability only from above (\cref{thm:pooled}). We also prove that the gate decouples learning the classes from learning the novelty (\cref{prop:decouple}).
\item With $90\%$ coarse labels, Arbor recognises $76$--$85\%$ of novel-class images on Cars, CUB and Aircraft, against at most $17\%$ for seven hierarchical classifiers, and $68\%$ on iNaturalist-2019 against at most $46\%$, at known-class accuracy within $1.6$ points of the best baseline on the first three (\cref{tab:merged,fig:curves}). The ranking by novel-class accuracy holds under four further label protocols, on the CUB--NABirds merge and against the released code of two prior methods (\cref{sec:results,app:official}).
\end{itemize}

\section{Related Work}\label{sec:related}

\paragraph{Hierarchical classifiers.} DARTS~\cite{deng2012hedging}, hierarchical selective classification~\cite{goren2024hsc} and CRM~\cite{karthik2021crm} read a node off flat leaf probabilities after training, and Valmadre~\cite{valmadre2022hierarchical} compares such rules across operating points and proposes a softmax over all nodes. Others weight leaf errors by tree distance~\cite{bertinetto2020making}, train node-conditional or per-level classifiers with a fallback~\cite{wu2020deeprtc,garg2022haf,lu2025bilt,park2025hcast,chu2026hieruq}, or put the hierarchy in the representation~\cite{sarkar2011tree,nickel2017poincare,ganea2018cones,khrulkov2020hyperbolic,snell2017prototypical,mettes2019hyperspherical,zeng2024hypstructure} (\cref{app:data}). None of these methods models why a label is coarse. The closest in method is hierarchical novelty detection~\cite{lee2018hierarchical}, which keeps a virtual novel class inside the same softmax in every superclass, whether or not any class there lacks fine-grained labels.

\paragraph{Labels at mixed granularity.} Merged datasets with labels at several granularities appear in detection~\cite{redmon2017yolo9000} and segmentation~\cite{bevandic2022multidomain}, FALCON~\cite{grcic2024falcon} discovers fine-grained classes from coarse labels, and many classifiers accept coarse labels directly~\cite{valmadre2022hierarchical,ristin2015coarse,taherkhani2019weakly,chen2022hmc,wang2026halo}, all without novel classes. Free-grained hierarchical visual recognition~\cite{park2026freegrain} is closest in setting: it trains per-level classifiers on the levels a label provides, completed with text and pseudo-labels, and returns a coarse label when uncertain. Our per-level baseline (\cref{sec:setup}) uses the same classifiers, and \cref{app:official} runs the released code of free-grained recognition.

\paragraph{Positive-unlabelled learning and loss correction.} Within a superclass, coarse labels pose a positive-unlabelled (PU) problem with several positive classes~\cite{xu2017mpu}. Arbor pools one coarsening probability over the known classes (\cref{sec:finite}), which assumes that labelled positives are selected completely at random~\cite{elkan2008pu,bekker2020pusurvey}. The identity of Elkan and Noto~\cite{elkan2008pu} then recovers the posterior of each known class. PU methods estimate the label frequency or mixture proportion~\cite{bekker2018tice,ramaswamy2016mpe,garg2021pulse} and correct the risk~\cite{duplessis2014pu,kiryo2017nnpu} in settings where negatives, here the novel classes, are known to exist. Most superclasses of a taxonomy contain none, and a correction applied there would invent novelty, so Arbor corrects only the superclasses in which a class has no fine-grained label. Our setting is also an instance of open-set label shift, with the images that carry fine-grained labels as source and the coarse-labelled ones as target. PULSE~\cite{garg2022osls} pairs a PU discriminator with a classifier learned from the labelled source and corrects the class priors. Arbor's classifier, unlike PULSE's, also learns from the coarse labels (\cref{prop:decouple}).

\paragraph{Coarse, partial and open-set labels.} Partial-label learning treats a coarse label as a candidate set, again under coarsening at random~\cite{cour2011partial,lv2020proden,wang2022pico}, and cannot tell a class from a sibling that is always labelled together with it~\cite{cour2011partial,liu2014learnability}. Coarse-to-fine learning~\cite{bukchin2021ancor,touvron2021grafit,chang2021flamingo,zhao2025twofold}, category discovery~\cite{vaze2022gcd,wen2023simgcd,cao2022orca,saito2021openmatch} and open-set recognition~\cite{bendale2016openmax,dhamija2018agnostophobia,hendrycks2019oe,vaze2022openset,sun2022knnood} treat novel classes as unlabelled or absent. Labelling that depends on the image~\cite{bekker2019sar,kato2019pubias} is outside Arbor's model.

\section{Method}\label{sec:methodall}
\subsection{What Coarse Labels Identify}\label{sec:theory}
A taxonomy is a rooted tree with nodes $\mathcal N$ and leaves $\mathcal C\subset\mathcal N$. We write $\ell\preceq y$ when leaf $\ell$ lies under node $y$, and call the parent of a leaf its \emph{superclass}. The classifier assigns to an image $x$ a probability $p_n=P(n\mid x)$ for every node, as in Valmadre's softmax over all nodes~\cite{valmadre2022hierarchical}.

\paragraph{The coarsening model.} We assume that an image of class $\ell$ is labelled with its superclass $y$ instead of $\ell$ with probability $q_\ell$, its \emph{coarsening probability}. We also assume that $q_\ell$ does not depend on the image given its class. A \emph{known} class has $q_\ell<1$ and a \emph{novel} class has $q_\ell=1$. The taxonomy lists every class, novel ones included, but no image shows what a novel class looks like, so the target for its images is the superclass. A class missing from the taxonomy is unseen, not novel (\cref{app:unseen}). With $p^*$ the true class posterior, the observed labels have likelihood
\begin{equation}
\begin{aligned}
P(\text{label }y\mid x)&=\textstyle\sum_{\ell\preceq y}q_\ell\,p^*(\ell\mid x),\\
P(\text{label }\ell\mid x)&=(1-q_\ell)\,p^*(\ell\mid x).
\end{aligned}
\label{eq:coarsen}
\end{equation}

\paragraph{Coarsening at random.} Write $Q^*_y(x)=\sum_{\ell\preceq y}p^*(\ell\mid x)$ for the total probability under $y$. The coarse term of \cref{eq:coarsen} factorises as
\begin{equation}
P(\text{label }y\mid x)=Q^*_y(x)\;\mathbb E_{\ell\sim p^*}\big[q_\ell\mid\ell\preceq y\big].
\label{eq:expect}
\end{equation}
The expectation in \cref{eq:expect} is the same for every class posterior if and only if $q_\ell$ is constant within the superclass (\cref{app:theory}). The labels are then \emph{coarsened at random}~\cite{heitjan1991ignorability,gill1997coarsening}, and the partial-label likelihood $Q_y=\sum_{\ell\preceq y}p_\ell$, the standard objective of the introduction, is exact. A novel class, with $q_\ell=1$ among siblings with $q_\ell<1$, violates coarsening at random.

\paragraph{Identification.} Write $\mathcal K_y$ and $\mathcal U_y$ for the known and novel classes of a superclass $y$. Given the coarsening probabilities, \cref{eq:coarsen} inverts for every known class: $p^*(\ell\mid x)=P(\text{label }\ell\mid x)/(1-q_\ell)$. This is the Elkan--Noto correction of positive-unlabelled learning~\cite{elkan2008pu}: a classifier trained on the observed labels, rescaled by the label frequency $1-q_\ell$. For the novel classes $\mathcal U_y$, \cref{eq:coarsen} gives only their total $p_{\mathcal U}(x)=\sum_{\ell'\in\mathcal U_y}p^*(\ell'\mid x)$~\cite{liu2014learnability}, since no label ever names one of them. Arbor therefore gives each such superclass a single virtual novel class, carrying $p_{\mathcal U}$.

\subsection{The Coarsening Likelihood}\label{sec:rule}
In the classifier the virtual novel class of $y$ is the node $y$ itself, so a prediction of $y$ reads ``$y$, unknown type''. A coarse label $y$ can then arise in two ways: from the virtual novel class, with probability $1$, or from a known class $\ell\preceq y$, with probability $q_\ell$. Arbor's loss for a coarse label is therefore
\begin{equation}
\boxed{\;\mathcal L_y(x)=-\log\Big(p_y+\sum_{\ell\preceq y}w_\ell\,p_\ell\Big)\;}
\label{eq:rule}
\end{equation}
with $p_y$ the probability of the virtual novel class and $w_\ell$ a weight that stands in for the unknown $q_\ell$ (\cref{sec:finite}). A fine-grained label keeps the log-loss $-\log p_\ell$. The partial-label likelihood drops $p_y$ and sets every $w_\ell=1$. Hierarchical novelty detection applied to coarse labels sets $w_\ell=0$. Setting $w_\ell=q_\ell$ gives \cref{eq:coarsen} itself, a forward loss correction~\cite{patrini2017losscorrection}. For a novel-class image, the partial-label likelihood is the same for every split of $Q_y$ below $y$, whereas \cref{eq:rule} with $w_\ell<1$ is minimised only by $p_y=1$ (\cref{app:theory}).

\subsection{The Optimum of the Loss}\label{sec:finite}
Two things remain: the weights $w_\ell$ and the population optimum of the loss. The labels alone cannot supply the $q_\ell$ (\cref{thm:pooled}(c) below). Counts can, if the classes within a superclass have equal size: a class then holds an equal share of its superclass's images, and the fraction of that share that kept a fine-grained label estimates $1-q_\ell$. On the fine-grained benchmarks this share rests on four to seven labels per class, so Arbor pools it into one estimate $\hat q$ over the known classes of the dataset (\cref{app:theory}) and sets, in every superclass with a virtual novel class,
\begin{equation}
w_\ell=\hat q\quad\text{for every class }\ell\preceq y.
\label{eq:weights}
\end{equation}
\paragraph{Where the virtual novel class is used.} Arbor places a virtual novel class in a superclass when one of its classes received no fine-grained label. Other superclasses keep the partial-label likelihood. A known class with $N$ training images is mistaken for novel only with probability $q_\ell^{\,N}$ (\cref{app:theory}).

The population optimum of this loss is explicit.
\begin{theorem}[The optimum of Arbor's loss]\label{thm:pooled}
Take an image whose class posterior has all its mass under a superclass $y$ with a virtual novel class, with observed-label probabilities $a_\ell(x)=P(\text{label }\ell\mid x)$ and $c(x)=P(\text{label }y\mid x)$. Over all node posteriors, the expected loss is minimised as follows (proof in \cref{app:theory}).
\begin{enumerate}[label=\textup{(\alph*)},leftmargin=1.6em,topsep=2pt,itemsep=1pt,parsep=0pt]
\item \textup{\textbf{Equivalence.}} Under any coarsening mechanism, $p_\ell\propto a_\ell$ and $p_y=\max\{0,\,(c-\hat q)/(1-\hat q)\}$, i.e.\ the Elkan--Noto correction applied to a classifier trained on the observed labels.
\item \textup{\textbf{Class-dependent coarsening.}} If $a_\ell=(1-q_\ell)p^*_\ell$ for the known classes, with $p^*_\ell=p^*(\ell\mid x)$, and the novel classes carry $p_{\mathcal U}$, then the ratios $p_\ell/p_{\ell'}$ are those of the partial-label likelihood and $p_y=\max\{0,\,p_{\mathcal U}+\sum_{\ell\in\mathcal K_y}p^*_\ell(q_\ell-\hat q)/(1-\hat q)\}$. With weights $w_\ell=q_\ell$ on the known classes instead, the optimum is exact: $p_\ell=p^*_\ell$ and $p_y=p_{\mathcal U}$.
\item \textup{\textbf{Identification.}} Any coarsening probabilities $q'_\ell\le q_\ell$ generate the same labels, with mass moved from the known classes to the novel ones, so the labels bound each $q_\ell$ only from above; the bound is reached on images certainly of class $\ell$, where $c=q_\ell$.
\end{enumerate}
\end{theorem}
\Cref{thm:pooled}(a) says that whatever the true coarsening, the optimal $p_y$ ranks images as the Elkan--Noto correction would, so any advantage of Arbor over the correction must come from training rather than from the optimum. Part (c) says that the labels rule out coarsening probabilities above the observed share of coarse labels but none below it.

\subsection{The Arbor Classifier}\label{sec:method}
\begin{table*}[t]\centering\scriptsize\setlength{\tabcolsep}{2.4pt}
\begin{tabular}{lcccccccccc}
\toprule
 & \multicolumn{2}{c}{Cars} & \multicolumn{2}{c}{CUB} & \multicolumn{2}{c}{CIFAR-100} & \multicolumn{2}{c}{Aircraft} & \multicolumn{2}{c}{iNat-2019} \\
\cmidrule(lr){2-3}\cmidrule(lr){4-5}\cmidrule(lr){6-7}\cmidrule(lr){8-9}\cmidrule(lr){10-11}
method & novel$\uparrow$ & known$\uparrow$ & novel$\uparrow$ & known$\uparrow$ & novel$\uparrow$ & known$\uparrow$ & novel$\uparrow$ & known$\uparrow$ & novel$\uparrow$ & known$\uparrow$ \\
\midrule
Flat softmax + HSC~\cite{goren2024hsc} & 6 & .844 & 12 & .715 & 17 & .888 & 7 & \textbf{.791} & 34 & .639 \\
Conditional softmax~\cite{redmon2017yolo9000,valmadre2022hierarchical} & 6 & .814 & 14 & .745 & \textbf{81} & .876 & 8 & .768 & 29 & .622 \\
Per-node sigmoids~\cite{valmadre2022hierarchical} & 14 & .826 & 17 & .757 & 56 & .893 & 17 & .762 & 28 & .598 \\
Per-level + fallback~\cite{park2025hcast,garg2022haf,chu2026hieruq} & 6 & .839 & 10 & \textbf{.759} & 34 & \textbf{.908} & 6 & .780 & 30 & \textbf{.646} \\
Hierarchical CE~\cite{bertinetto2020making} & 10 & .804 & 14 & .707 & 36 & .873 & 11 & .762 & 30 & .607 \\
Soft labels~\cite{bertinetto2020making} & 6 & \textbf{.851} & 12 & .724 & 32 & .884 & 8 & .784 & 29 & .637 \\
HypStructure~\cite{zeng2024hypstructure} & 8 & .834 & 11 & .713 & 23 & .887 & 7 & .767 & 35 & .616 \\
\midrule
Uniform target & 39 & .799 & 24 & .637 & 67 & .753 & 42 & .784 & 46 & .467 \\
Hierarchical novelty detection~\cite{lee2018hierarchical} & 68 & .738 & 45 & .637 & 78 & .816 & 57 & .708 & 43 & .441 \\
\midrule
\multicolumn{11}{@{}l}{\textit{post-hoc corrections}} \\
Elkan--Noto correction~\cite{elkan2008pu} & 77 & .738 & 68 & .637 & 78 & .816 & 78 & .708 & -- & -- \\
Two-stage (best classifier + Elkan--Noto) & 66 & .839 & 72 & .745 & 75 & \textbf{.908} & 76 & \textbf{.791} & -- & -- \\
\midrule
\rowcolor{ourspastel} \textbf{Arbor (ours)} & \textbf{85} & .846 & \textbf{76} & .742 & 80 & .876 & \textbf{83} & .784 & \textbf{68} & .594 \\
\bottomrule
\end{tabular}

\caption{\textbf{Mixed-granularity labels: $90\%$ coarse labels and one novel class per superclass.} Novel: novel-class accuracy (\%). Known: known-class accuracy (leaf posterior). Novel-class accuracy at the same operating point (\cref{sec:setup}), known-class accuracy from the leaf argmax, three seeds. Seed-level differences and leaf shares: \cref{tab:op,tab:merged_full}. The Elkan--Noto correction and the two-stage baseline were not run on iNat-2019. Bold: best in column, including ties within half a point or at the precision shown.}
\label{tab:merged}\end{table*}
The analysis so far treated the node probabilities $p$ as free. Arbor parameterises them with a softmax $r$ over all nodes, the leaves of novel classes included, and a novelty gate whose output $P$ enters the loss. \Cref{thm:pooled}(a) gives the novel leaves no mass at the optimum, since no label ever names them. A frozen backbone~\cite{oquab2024dinov2} and a two-layer projector map an image to $z\in\R^{k}$. Each node $n$ has a prototype $\phi_n\in\R^k$. The root sits at the origin. Each child sits at a distance from its parent $\mathrm{pa}(n)$ that is shared by all nodes of its depth,
\begin{equation}
\phi_n=\phi_{\mathrm{pa}(n)}+L_{\mathrm{depth}(n)}\,u_n/\|u_n\|,
\label{eq:proto}
\end{equation}
with a learned direction $u_n\in\R^k$ and one learned length $L_d>0$ per depth $d$. One softmax runs over all nodes, with a learned temperature $\tau>0$ and a bias $b_n\in\R$ per node,
\begin{equation}
r(n\mid x)\propto\exp\big(-\|z-\phi_n\|^2/\tau+b_n\big).
\label{eq:softmax}
\end{equation}
\paragraph{The novelty gate.} The virtual novel class is still the node $y$ of the final posterior $P$ (\cref{sec:rule}), but its probability is not the softmax term $r(y\mid x)$. Arbor gives each superclass $y$ with a virtual novel class, a \emph{gated superclass}, a logistic gate $\nu_y(x)=\sigma(g_y^\top z+\beta_y)$, with parameters $g_y\in\R^k$, $\beta_y\in\R$. The gate decides what fraction of the superclass's softmax mass $m_y(x)=\sum_{\ell\preceq y}r(\ell\mid x)$ is novel:
\begin{equation}
\begin{aligned}
P(y\mid x)&=\nu_y\,m_y,\\
P(\ell\mid x)&=(1-\nu_y)\,r(\ell\mid x)\quad\text{for }\ell\preceq y,
\end{aligned}
\label{eq:gate}
\end{equation}
and every other node keeps $P(n\mid x)=r(n\mid x)$. The softmax term $r(y\mid x)$ of $y$ itself is unused, and reported probabilities are renormalised (\cref{app:lemmas}). The loss is \cref{eq:rule} on $P$ for a coarse label and $-\log P(\ell\mid x)$ for a fine-grained one. With a penalty on the embedding of weight $\rho$, the loss reads
\begin{equation}
\mathcal L(x,t)=-\log\sum_{n\in\mathcal N}w_t(n)\,P(n\mid x)+\rho\,\|z\|^2,
\label{eq:loss}
\end{equation}
where $t$ is the observed label and $w_t$ collects the weights. For a fine-grained label, $w_t(n)=\mathbf 1[n=t]$. For a coarse label $y$ with a virtual novel class, $w_t(y)=1$ and $w_t(\ell)=\hat q$ on $\ell\preceq y$ (\cref{eq:rule,eq:weights}). For a coarse label $y$ without one, $w_t(\ell)=1$ on $\ell\preceq y$. On every other node, $w_t(n)=0$.
\begin{proposition}[Decoupling]\label{prop:decouple}
Under \cref{eq:gate}, Arbor's loss without the embedding penalty is the sum of two losses that share no parameter except those of the embedding $z$:
\[\small\renewcommand{\arraystretch}{1.25}\setlength{\arraycolsep}{2.5pt}\begin{array}{lcc}
 & \text{fine-grained label }\ell & \text{coarse label }y\\\hline
\text{classes } (r) & -\log r(\ell\mid x) & -\log m_y(x)\\
\text{novelty } (\nu_y) & -\log(1-\nu_y) & -\log(\hat q+(1-\hat q)\nu_y)
\end{array}\]
The first row is the partial-label likelihood of the softmax $r$; the second is the positive-unlabelled likelihood of the gate, with label frequency $1-\hat q$.
\end{proposition}
\Cref{prop:decouple} shows in particular that $\hat q$ enters only the gate.
The proof is a one-line identity (\cref{app:lemmas}), and \cref{thm:pooled} applies to the gated model unchanged. With the virtual novel class as a node of the softmax instead~\cite{lee2018hierarchical}, the two terms are coupled: a coarse label flattens the known siblings by an amount that grows with the image's novelty mass (\cref{prop:em}).

\paragraph{Prediction.} The prediction is the most probable node under a depth prior, with $\lambda$ set on validation (\cref{sec:setup}):
\begin{equation}
\hat n=\arg\max_{n}\;\log P(n\mid x)+\lambda\,\mathrm{depth}(n).
\label{eq:readout}
\end{equation}

\section{Experiments}\label{sec:experiments}
\begin{figure*}[t]\centering\includegraphics[width=\textwidth]{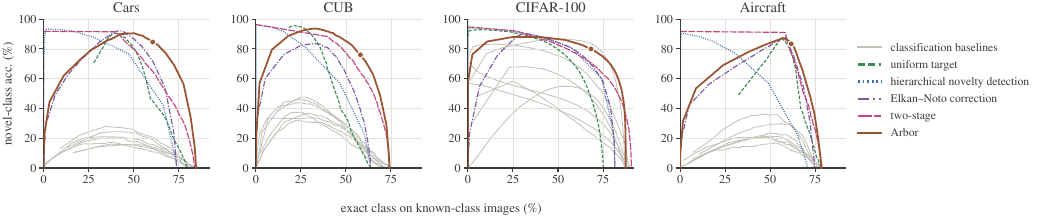}
\caption{\textbf{Novel-class accuracy across operating points,} against the exact-class rate on known-class images, main protocol, three seeds; dot: Arbor's operating point. Arbor's curve is the highest over most of the range on Cars, CUB and Aircraft and over two stretches on CIFAR-100 (intervals in \cref{app:op}).}\label{fig:curves}\end{figure*}
\subsection{Setup}\label{sec:setup}
\paragraph{Data.} We use Stanford Cars~\cite{krause2013cars} (196 classes), CUB-200-2011~\cite{wah2011cub} (200), CIFAR-100~\cite{krizhevsky2009cifar} (100) and FGVC-Aircraft~\cite{maji2013aircraft} (100), each with a three-level taxonomy (constructed from the class names for Cars and CUB, \cref{app:data}), and iNaturalist-2019~\cite{vanhorn2018inat} (iNat-2019). All methods train a classifier on frozen DINOv2-B/14 features~\cite{oquab2024dinov2}, and every result averages three seeds unless stated otherwise (sizes, splits and embedding dimensions in \cref{app:data}). Under the main protocol below, with $90\%$ coarse labels, a class keeps about $42$ fine-grained labels on CIFAR-100, four to seven on Cars, CUB and Aircraft, and about $19$ on average on iNat-2019.

\paragraph{Protocols.} \emph{Mixed granularity} (the main protocol): in every superclass of two or more classes, the first listed class is novel. Of all other training and validation images, $90\%$ carry only their superclass label. \emph{Two novel classes}, \emph{$50\%$ coarse labels}, \emph{a random novel class} and \emph{class-dependent coarsening} ($q_\ell\sim U[0.80,0.98]$) each change one setting of the main protocol (\cref{app:moreprotocols,app:varq}). \emph{Special cases}: all labels fine-grained, $90\%$ random coarse labels with every class known, and only the novel classes coarse.

\paragraph{Metrics.} Let $\mathcal D_{\mathrm K}$ and $\mathcal D_{\mathrm U}$ be the known- and novel-class test images. Let $\ell(x)$ be the class of $x$, $\mathrm{pa}(\ell)$ its superclass and $\hat n(x)$ a method's predicted node (Arbor: \cref{eq:readout}). Every class sits at depth $D=3$. Known-class accuracy reads the most probable class and depends on no threshold. Novel-class accuracy~\cite{lee2018hierarchical} reads the prediction at the \emph{operating point}, the threshold setting defined at the end of this paragraph, and counts a novel-class image as correct when that prediction is its superclass (formulas in \cref{app:op}). The exact-class rate $\mathrm{ex}$ is the share of known-class images whose prediction at the operating point is their exact class. The incorrect rate $\mathrm{inc}$ is the share of all test images whose prediction is neither their class nor an ancestor of it (\cref{app:data}). Each method's threshold ($\lambda$ in \cref{eq:readout}, or a confidence threshold) defines its operating point. We set it so that the mean specificity $\mathrm{depth}(\hat n(x))/D$ over validation images with fine-grained labels is $0.9$. The tables compare methods at that operating point, and \cref{fig:curves} compares them across all operating points~\cite{valmadre2022hierarchical}. Differences quoted in the text are computed from unrounded values.
\begin{table}[t]\centering\scriptsize\setlength{\tabcolsep}{3pt}
\begin{tabular}{lcccc}
\toprule
 & Cars & CUB & CIFAR-100 & Aircraft \\
\midrule
\multicolumn{5}{@{}l}{\textit{exact-class rate on known-class images (\%)}} \\
best classification baseline & \textbf{77} & \textbf{63} & \textbf{78} & \textbf{73} \\
best novelty baseline & 64 & 54 & 61 & 68 \\
\rowcolor{ourspastel} \textbf{Arbor} & 61 & 58 & 68 & 62 \\
\midrule
\multicolumn{5}{@{}l}{\textit{incorrect predictions over all test images (\%)}} \\
best classification baseline & 22 & 23 & \textbf{6} & 23 \\
best novelty baseline & 17 & 24 & 13 & 14 \\
\rowcolor{ourspastel} \textbf{Arbor} & \textbf{11} & \textbf{16} & 10 & \textbf{12} \\
\midrule
\multicolumn{5}{@{}l}{\textit{novel-class accuracy at the baseline's exact-class rate (\%)}} \\
best classification baseline & 14 & 17 & \textbf{81} & 17 \\
\rowcolor{ourspastel} \textbf{Arbor}, matched & \textbf{70} & \textbf{68} & 76 & \textbf{70} \\
\bottomrule
\end{tabular}

\caption{\textbf{Predictions at the same operating point}, main protocol, three seeds. Best classification / novelty baseline: that group's best member per row; bottom block: Arbor matched on validation to the exact-class rate of the classification baseline that recognises the most novel classes. Bold: best in column within a block (\cref{tab:op}).}
\label{tab:deployed}\end{table}
\begin{table}[t]\centering\scriptsize\setlength{\tabcolsep}{3pt}
\begin{tabular}{lcccc}
\toprule
 & Cars & CUB & CIFAR-100 & Aircraft \\
\midrule
\multicolumn{5}{@{}l}{\textit{novel-class accuracy (\%)}} \\
Elkan--Noto correction & 77 & 68 & 78 & 78 \\
two-stage & 66 & 72 & 75 & 76 \\
\rowcolor{ourspastel} \textbf{Arbor} & 85 & 76 & 80 & \textbf{83} \\
\rowcolor{ourspastel} \textbf{Arbor}, matched to Elkan--Noto & \textbf{87} & \textbf{86} & \textbf{82} & \textbf{83} \\
\midrule
\multicolumn{5}{@{}l}{\textit{incorrect predictions, all test images (\%)}} \\
Elkan--Noto correction & 15.4 & 22.5 & 13.8 & 13.3 \\
two-stage & 15.1 & 18.6 & 11.0 & 12.7 \\
\rowcolor{ourspastel} \textbf{Arbor} & 10.7 & 16.2 & 10.4 & 12.3 \\
\rowcolor{ourspastel} \textbf{Arbor}, matched to Elkan--Noto & \textbf{9.6} & \textbf{12.1} & \textbf{9.4} & \textbf{12.2} \\
\bottomrule
\end{tabular}

\caption{\textbf{The Elkan--Noto correction and the two-stage baseline} against Arbor, main protocol, three seeds: novel-class accuracy and incorrect predictions (over all test images, \%) at each method's own operating point, and in the last row of each block with Arbor's threshold set on validation to the Elkan--Noto correction's exact-class rate. Known-class accuracies: \cref{tab:merged}; full table: \cref{tab:pu_full}. Bold: best in column within a block.}
\label{tab:pu}\end{table}
\paragraph{Baselines.} Classification: the seven hierarchical classifiers of \cref{sec:related} listed in \cref{tab:merged} (provenance in \cref{app:data}; \cref{app:official} runs released implementations) and, in the special cases, five-member ensembles~\cite{lakshminarayanan2017ensembles}, hyperbolic Busemann~\cite{atigh2021busemann} and a flat softmax with a min/max weighting of its coarse labels (\cref{app:data}). Novelty: the uniform target (a uniform distribution over a coarse label's classes~\cite{cour2011partial}; \cref{prop:maxent}), and hierarchical novelty detection~\cite{lee2018hierarchical}. Post hoc: the Elkan--Noto correction~\cite{elkan2008pu} applied to the hierarchical novelty detection classifier, and a two-stage baseline that pairs the most accurate classification baseline with the correction's novelty (\cref{app:pu}). All share the backbone and projector (\cref{tab:baselines_mech}).
\paragraph{Selection.} Each method chooses its variant, embedding penalty, classifier learning rate and operating point on validation (grids and exceptions in \cref{app:data}; \cref{prop:radius} motivates the penalty grid of the classifiers with a virtual novel class). Three design choices were first made on test results: the coarse-label weights of \cref{eq:rule,eq:weights}, the novelty gate and the exclusion of $\rho=0$. Redone on validation with pseudo-novel classes, known classes that keep only their superclass label, the selection rule picks the gate with pooled $\hat q$ over every candidate and never picks $\rho=0$, but among single-softmax weightings it ranks a min/max weighting above the pooled one (\cref{app:data}, \cref{tab:pseudonovel}). The gate's adoption on the six further settings followed a criterion fixed before those runs, and the naturally coarse iNaturalist labels, the iNat-2019 special cases, the unseen-class protocol, the few-fine-labels protocol and the CUB official split were not used for any of the three choices (\cref{app:data}). Matched comparisons set Arbor's threshold to a baseline's exact-class rate on validation, except on the merge and in \cref{fig:curves}, where it is read from the test curves.

\subsection{Results}\label{sec:results}
\begin{figure*}[t]\centering\includegraphics[width=\textwidth]{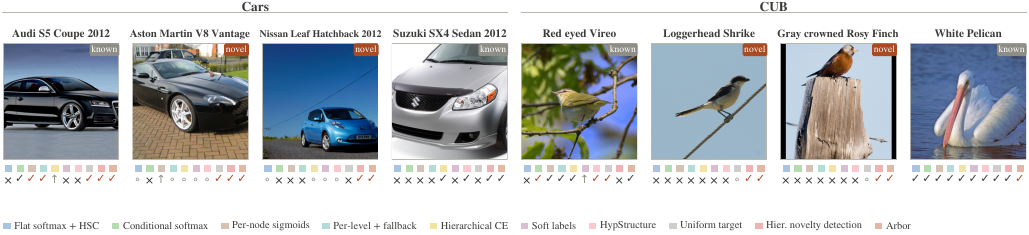}
\caption{\textbf{Test images drawn at random} (fixed seed, distinct classes; main protocol; four from Cars and four from CUB, half of novel and half of known classes; three benchmarks in \cref{fig:examples_full}). Under each image, one mark per method, legend order, Arbor last, at the shared operating point. Black $\checkmark$: the exact class. Orange $\checkmark$: the superclass, correct for a novel class and coarse for a known one. $\times$: a wrong class. $\circ$: a novel class's own leaf. $\uparrow$: an ancestor above the superclass.}
\label{fig:examples}\end{figure*}
\begin{figure}[t]\centering\includegraphics[width=\columnwidth]{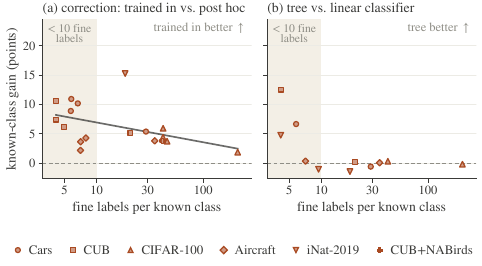}
\caption{\textbf{Arbor's gain in known-class accuracy grows as fine-grained labels become scarce.} Each point is one benchmark under one protocol; $x$: fine-grained labels per known class; $y$: Arbor's known-class accuracy minus that of (a) the Elkan--Noto correction (equal at the optimum; line: least-squares fit) and (b) a linear classifier with the same loss and gate. Three seeds (one seed for the linear classifier on iNat-2019 at $90\%$ coarse labels; \cref{app:pu}). For iNat-2019, panel (a) uses the known-class accuracy of hierarchical novelty detection, since the correction was not run there.}\label{fig:gains}\end{figure}
\paragraph{Main protocol.} On Cars, CUB and Aircraft, Arbor recognises $76$--$85\%$ of novel-class images with known-class accuracy within $1.6$ points of the most accurate classification baseline (\cref{tab:merged,fig:examples}). The novelty baselines recognise novel classes at a cost in known-class accuracy. The uniform target, for example, pulls siblings towards equal probability (\cref{prop:maxent}). With its threshold matched to a classification baseline's exact-class rate, Arbor recognises four to five times as many novel-class images as that baseline on these three benchmarks (\cref{tab:deployed}, bottom).

On CIFAR-100, Arbor is $3.3$ points less accurate on known classes. The cost comes through the embedding shared with the gate (\cref{prop:decouple}). Elsewhere it is offset by learning the known classes from coarse-labelled images (\cref{fig:gains}a). With about $42$ fine-grained labels per class, that gain is smallest on CIFAR-100. The conditional softmax's $1$-point advantage in novel-class accuracy on CIFAR-100 does not survive a random choice of novel class, $81$ to $51\%$ (\cref{app:moreprotocols}). Even when told at inference which superclasses contain a novel class, the baselines remain below Arbor in $98$ of $108$ seed-level comparisons (\cref{app:countaware}).

\paragraph{Exact-class rate.} At the same operating point, Arbor's exact-class rate on known-class images is $4.6$--$16.5$ points below the highest rate among the classification baselines: the baselines spend their superclass answers on their least confident images in every superclass, Arbor mainly in the gated superclasses (\cref{app:op}). Over all test images, novel-class images included, Arbor therefore makes the fewest incorrect predictions on Cars, CUB and Aircraft. On CIFAR-100 the best classification baseline makes fewer (\cref{tab:deployed}).

\paragraph{Elkan--Noto correction and two-stage baseline.} By \cref{thm:pooled}(a) the Elkan--Noto correction equals Arbor at the optimum, so this comparison tests whether training the correction into the classifier matters (\cref{tab:pu}). It does: the correction is $2$--$8$ points below Arbor on novel-class accuracy, and matched to the correction's exact-class rate, Arbor recognises more novel-class images and makes fewer incorrect predictions on every benchmark. The corrected classifier learns the known classes from fine-grained labels only (\cref{tab:merged}). Arbor's softmax also learns them from every coarse-labelled image, for any $\hat q$ (\cref{prop:decouple}). Across $18$ benchmark--protocol settings (\cref{fig:gains}a), Arbor's advantage on known classes grows as fine-grained labels per class fall. The two-stage baseline recognises fewer novel-class images on every benchmark, although on CUB, CIFAR-100 and Aircraft its classifier is the more accurate.

\paragraph{Other protocols and iNat-2019.} Under the other label protocols, Arbor remains above the classification baselines in novel-class accuracy at a cost of at most $5.3$ points of known-class accuracy (\cref{app:moreprotocols,app:varq}). With class-dependent coarsening, a pooled $\hat q$ sends heavily coarsened known classes to their superclass, as \cref{thm:pooled}(b) predicts: the most-coarsened third of known classes are predicted as their superclass far more often than the least-coarsened third, $68$--$80$ vs $7$--$43\%$ (\cref{app:varq}). On iNat-2019, Arbor recognises $68\%$ of novel-class images against $46\%$ for the uniform target, the best baseline. It is $5.3$ points below per-level + fallback on known classes and makes $5.4$ points more incorrect predictions than the flat softmax. The known-class gap is due to the single softmax over all nodes rather than to the virtual novel class, which accounts for only $1$ point of it (\cref{app:inat}).

\paragraph{CUB--NABirds merge.} In the CUB--NABirds merge~\cite{vanhorn2015nabirds} (\cref{app:merge}), each known class is coarsened with probability equal to its NABirds share, $0.34$--$0.88$. Arbor recognises $94.1\%$ of novel species against $93.2\%$ for the next-best baseline, and more than each of six baselines when its threshold is matched to that baseline's exact-class rate (\cref{tab:merge}). Its novelty score matches the Elkan--Noto correction's to within $0.01$ of AUROC, as \cref{thm:pooled}(a) predicts. Every classification baseline is $2.4$--$4.4$ points more accurate on known classes. With about $42$ fine-grained labels per known class, the training advantage of \cref{fig:gains}a is small. The two-stage baseline, the most accurate, also makes $0.5$ points fewer incorrect predictions. It recognises fewer novel species, $92.6\%$ (\cref{app:merge}).

\paragraph{Special cases.} With every class known the partial-label likelihood is exact and Arbor places almost no virtual novel classes (\cref{sec:theory,app:theory}). Its known-class accuracy is then within $1.8$ points of the best single model on the four small benchmarks and $3.5$--$4.4$ points below it on iNat-2019, and with only the novel classes coarse it recognises $78$--$88\%$ of novel-class images (\cref{tab:main}, \cref{app:full}).

\begin{table}[!tb]\centering\scriptsize\setlength{\tabcolsep}{1.5pt}
\begin{tabular}{lcccccccc}
\toprule
 & \multicolumn{2}{c}{Cars} & \multicolumn{2}{c}{CUB} & \multicolumn{2}{c}{CIFAR-100} & \multicolumn{2}{c}{Aircraft} \\
\cmidrule(lr){2-3}\cmidrule(lr){4-5}\cmidrule(lr){6-7}\cmidrule(lr){8-9}
variant & K$\uparrow$ & N$\uparrow$ & K$\uparrow$ & N$\uparrow$ & K$\uparrow$ & N$\uparrow$ & K$\uparrow$ & N$\uparrow$ \\
\midrule
\rowcolor{ourspastel} \textbf{Arbor} & .846 & 85 & .742 & 76 & .876 & 80 & .784 & 83 \\
no virtual novel class & .866 & 0 & .754 & 3 & .895 & 1 & .804 & 1 \\
gate everywhere & .843 & 74 & .734 & 74 & .876 & 80 & .774 & 53 \\
per-class $\hat q_\ell$ & .695 & 64 & .527 & 46 & .837 & 33 & .690 & 63 \\
$\hat q=0$ & .813 & 75 & .709 & 54 & .872 & 83 & .769 & 79 \\
\midrule
novelty in the softmax & .837 & 49 & .733 & 48 & .876 & 78 & .794 & 59 \\
novelty in the softmax, novel leaves at $0$ & .845 & 70 & .742 & 63 & .876 & 74 & .791 & 59 \\
fully factorised & .802 & 80 & .672 & 62 & .829 & 76 & .779 & 84 \\
\midrule
free prototypes & .827 & 83 & .729 & 76 & .877 & 82 & .772 & 83 \\
linear classifier & .779 & 82 & .681 & 43 & .872 & 72 & .780 & 82 \\
\bottomrule
\end{tabular}

\caption{\textbf{Ablation}, mixed-granularity labels; K / N: known- / novel-class accuracy; novel leaves at $0$: zero weight on the leaves of classes without fine-grained labels, known classes keeping $\hat q$; fully factorised: one softmax over the superclasses and one within each.}
\label{tab:abl}\end{table}
\subsection{Ablation}\label{sec:ablation}
\begin{itemize}[leftmargin=1.2em,itemsep=1pt,topsep=2pt]
\item \emph{Virtual novel class} (\cref{tab:abl}). Without it, the classifier recognises $0$--$3\%$ of novel-class images. With it, known-class accuracy is $1$--$2$ points lower. A gate in every superclass lowers novel-class accuracy by $0$--$31$ points: with no novelty to find, it turns the error in $\hat q$ into spurious novelty (\cref{thm:pooled}(b)).
\item \emph{The pooled weight $\hat q$.} Without it ($\hat q=0$) the gate must explain every coarse label, so by \cref{thm:pooled}(a) a known-class image receives novelty mass near its coarsening probability. Known-class accuracy falls by $0.4$--$3.3$ points and novel-class accuracy by $5$--$22$ on Cars, CUB and Aircraft. A separate estimate $\hat q_\ell$ per class rests on four to seven fine-grained labels per class on these benchmarks. Its error becomes misplaced novelty mass (\cref{thm:pooled}(b)), costing $9$--$22$ points of known-class accuracy.
\item \emph{Novelty parametrisation.} With the virtual novel class as a softmax node, the same loss has $23$ points lower novel-class accuracy on average, at known-class accuracy within $1$ point (\cref{prop:decouple}; \cref{fig:decouple}). In one softmax a coarse-labelled image flattens the known siblings (\cref{prop:em}), and the virtual novel class competes with them for the same coarse-label mass, which the known leaves, trained also by fine-grained labels, win. The gate removes this competition (\cref{prop:decouple}). A fully factorised classifier, whose coarse-labelled images do not train the classes of their superclass, loses $0.5$--$7$ points of known-class accuracy.
\item \emph{Classifier.} With the same loss and gate, linear logits are $0.3$--$6.7$ points less accurate on known classes and $1$--$33$ points lower in novel-class accuracy. Free prototypes cost at most $1.9$ points of known-class accuracy. The gain from the tree's inductive bias (\cref{sec:method}) is largest where fine-grained labels are fewest (\cref{fig:gains}b).
\end{itemize}

\section{Conclusion and Limitations}
A class that never receives a fine-grained label breaks coarsening at random, so the standard objective has no reason to find it. Arbor models the coarsening with a separate logistic gate, and recognises most such classes at a known-class accuracy close to the best classifier's. The advantage is largest where fine-grained labels are scarce and the coarsening is a property of the class rather than of the image.

\paragraph{Limitations.} Arbor assumes that the coarsening probability depends only on the class, not on the image, and that the taxonomy lists every class. On iNaturalist's own genus-level labels, which may arise when a photograph is hard, the gate may partly learn to detect hard photographs, and Arbor is below hierarchical novelty detection on novel species ($49$ vs $56\%$; \cref{app:natural}). One pooled $\hat q$ cannot tell unequal coarsening among known classes from novelty (\cref{thm:pooled}(b), \cref{app:varq}), and a PULSE-style correction per superclass~\cite{garg2022osls} remains untested. A superclass prediction does not certify novelty (\cref{app:op}). A class absent from the taxonomy, unseen rather than novel, is read as a sibling leaf (\cref{app:unseen}). A known class with one or two fine-grained labels is classified against siblings that keep all of theirs and goes to a sibling (\cref{prop:absorb}, \cref{app:weak}).

\section*{LLM Usage Statement}
Claude (Anthropic) was used as an assistant under the authors' direction. Code and experiments: it wrote the analysis and plotting scripts and ran the experiment pipeline. Figures: it drafted every figure except \cref{fig:overview}, which the authors made, from the saved results. Text: it polished the prose, checked the reference list and searched the literature for related work. It contributed no theoretical result and no research idea. The authors verified every output and take full responsibility for the content.

{\small\bibliographystyle{ieeenat_fullname}\bibliography{main}}
\clearpage
\appendix
\section{Coarsened labels: proofs and estimators}\label{app:theory}
Proofs for \cref{sec:theory,sec:rule,sec:finite}, and the count estimate of $\hat q$.

Throughout, consider one superclass $y$ with leaves $\mathcal C_y$, $K=|\mathcal C_y|$. An image of class $\ell$ is labelled $y$ with probability $q_\ell$ and $\ell$ otherwise, independently of $x$ given $\ell$. The known leaves are $\mathcal K_y=\{\ell:q_\ell<1\}$. The novel leaves, whose training images carry only the superclass label, are $\mathcal U_y=\mathcal C_y\setminus\mathcal K_y$.

\paragraph{Observed-data likelihood.} Summing over the unobserved class gives $P(\text{label}=y\mid x)=\sum_\ell P(\ell\mid x)\,P(\text{coarse}\mid\ell)=\sum_\ell q_\ell\,p^*(\ell\mid x)$ and $P(\text{label}=\ell\mid x)=(1-q_\ell)\,p^*(\ell\mid x)$, which is \cref{eq:coarsen}.

\paragraph{Coarsening at random (\cref{sec:theory}).} If $q_\ell\equiv q$, the coarse term is $q\,Q_y$, so $-\log P(y\mid x)=-\log Q_y-\log q$. Conversely, suppose $\sum_\ell q_\ell p_\ell=c_0\sum_\ell p_\ell$ for every $p$ in the simplex. Taking $p$ to be each vertex in turn gives $q_\ell=c_0$. The total probability $Q_y$ depends on the nodes below $y$ only through their sum, so any reallocation among them leaves it unchanged. With logits $o_n$ and $p_n=e^{o_n}/Z$, $\partial(-\log Q_y)/\partial o_m=p_m-\mathbf 1[m\preceq y]\,p_m/Q_y=-p_m(1-Q_y)/Q_y$ for $m\preceq y$. Every node under $y$ is therefore raised at a rate set by its current probability. A fine-grained label $\ell'$ outside $\{y\}$ contributes $\partial(-\log p_{\ell'})/\partial o_y=p_y$, which pushes $o_y$ down. $\square$

\paragraph{Identification (\cref{sec:theory}).} For $\ell\in\mathcal K_y$, $p^*(\ell\mid x)=P(\text{label}=\ell\mid x)/(1-q_\ell)$ is a function of the observed label distribution. Then $P(\text{label}=y\mid x)=\sum_{\ell\in\mathcal K_y}q_\ell\,p^*(\ell\mid x)+p_{\mathcal U}(x)$ gives $p_{\mathcal U}(x)$. No observed label involves a single novel leaf, because novel leaves carry no fine-grained label and share coarsening probability one. The observed distribution therefore depends on $\{p^*(\ell'\mid x)\}_{\ell'\in\mathcal U_y}$ only through their sum. For $|\mathcal U_y|\ge2$, any two splits of $p_{\mathcal U}$ are observationally equivalent. If $\mathcal K_y=\emptyset$, only $\sum_\ell p^*(\ell\mid x)$ is identified. This is the case of ambiguity degree one in \citet{liu2014learnability}. $\square$

\paragraph{The loss as the observed-data likelihood.} Let the virtual novel class carry $p_{\mathcal U}$, so $p_y:=p_{\mathcal U}$. Then $-\log P(\text{label}=y\mid x)=-\log(p_y+\sum_{\ell\in\mathcal K_y}q_\ell p_\ell)$. This is the loss \cref{eq:rule} with $w_y=1$ and $w_\ell=q_\ell$. The fine term $-\log P(\text{label}=\ell\mid x)=-\log p_\ell-\log(1-q_\ell)$ differs from the fine-label loss by a constant.

\paragraph{\Cref{thm:pooled} (the optimum of Arbor's loss).} Fix $x$ with its posterior within $y$. Write $a_\ell$ for the probability that its observed label is the fine-grained label $\ell$, and $c=1-\sum_\ell a_\ell$ for the probability that it is the coarse label $y$. With weights $w_\ell=\hat q$, the expected loss is $-\sum_\ell a_\ell\log p_\ell-c\log(p_y+\hat q S)$, with $S=\sum_\ell p_\ell$ over the known classes and $p_y=1-S$; a class without fine-grained labels carries no mass at the optimum, since moving its mass to $p_y$ leaves the fine term unchanged and lowers the coarse one. For fixed $S$, only the first term depends on how $S$ splits among the known classes, and it is minimised at $p_\ell\propto a_\ell$; the partial-label likelihood, whose coarse term $-c\log S$ does not depend on the split either, has the same split. With $A=\sum_\ell a_\ell=1-c$, the loss in $S$ is $-A\log S-c\log(1-(1-\hat q)S)$, which is convex with its minimum at $S^*=A/(1-\hat q)$. If $S^*\le1$, then $p_\ell=a_\ell/(1-\hat q)$ and $p_y=1-S^*=(c-\hat q)/(1-\hat q)$; otherwise $S=1$ and $p_y=0$. This is (a): the Elkan--Noto correction $p^*_\ell=a_\ell/(1-\hat q)$ with the residual on the virtual novel class, which uses only the observed-label probabilities and so holds under any coarsening mechanism.

For (b), substitute $a_\ell=(1-q_\ell)p^*_\ell$ and $c=\sum_\ell q_\ell p^*_\ell+p_{\mathcal U}$; the pooled formula follows from (a). With $w_\ell=q_\ell$ the expected loss is the cross-entropy between the observed label distribution and the one the model implies, which by Gibbs' inequality is minimised when they match; $p_\ell=p^*_\ell$ and $p_y=p_{\mathcal U}$ achieve this. The most probable known class is the Bayes class whenever $p^*_\ell/p^*_m>(1-q_m)/(1-q_\ell)$; the right-hand side is at most $(1-q_{\min})/(1-q_{\max})$, with $q_{\min}$ and $q_{\max}$ the smallest and largest $q_\ell$.

For (c), given $q'$ with $q'_\ell\le q_\ell$, set $p'_\ell=(1-q_\ell)p^*_\ell/(1-q'_\ell)\le p^*_\ell$ and $p'_{\mathcal U}=1-\sum_\ell p'_\ell\ge p_{\mathcal U}\ge0$; then $(1-q'_\ell)p'_\ell=a_\ell$ and $p'_{\mathcal U}+\sum_\ell q'_\ell p'_\ell=c$, so $(p',p'_{\mathcal U},q')$ generates the same labels. If $q'_\ell>q_\ell$, an image with $p^*_\ell=1$ would need $p'_\ell=(1-q_\ell)/(1-q'_\ell)>1$; on such an image $c=q_\ell$. $\square$

\paragraph{Over- versus underestimates of $\hat q$.} By (a), $p_y=(c-\hat q)/(1-\hat q)$ is increasing in $c$ for every $\hat q$, so $\hat q$ changes the ranking of images by $p_y$ only where the clip at zero binds. An overestimate makes the clip bind on every image with $c<\hat q$; these images are tied at zero and their ranking is lost. An underestimate changes $p_y$ by an increasing map of $c$, which setting the operating point on validation largely compensates, so an underestimate is the safer error. The held-out likelihood is flat in a pooled $\hat q$ up to the smallest coarsening probability among the known classes and falls above it. It can therefore reject an overestimate but not choose among underestimates, and so selects the smallest candidate $q$ on the merge and on naturally coarse labels.

\paragraph{Estimating the coarsening probability from counts.} Assume the classes within each superclass are balanced. Then every leaf of $y$ has $\bar N_y=(n^{\mathrm c}_y+\sum_\ell n_\ell)/K$ training images in expectation, where $n^{\mathrm c}_y$ counts the coarse-labelled ones. Hence $\mathbb E[n_\ell]=(1-q_\ell)\bar N_y$. Pooling the known leaves of all superclasses that carry coarse labels gives $\hat q=1-\sum_y\sum_{\ell\in\mathcal K_y}n_\ell\big/\sum_y|\mathcal K_y|\,\bar N_y$. This is $0.898$--$0.903$ on the four benchmarks at $q=0.9$, with and without novel classes. The single-class superclasses, in which every coarse label belongs to a known class, give $0.900$, $0.893$ and $0.896$ on Cars, CUB and Aircraft (three seeds). It is $0$ when the known classes keep all their fine-grained labels. A per-leaf estimate $1-n_\ell/\bar N_y$ relies on the four or so fine-labelled images a CUB class keeps at $q=0.9$. With about $42$ images per class, a binomial count misses it by more than $0.03$ for about $43\%$ of the classes, which by \cref{thm:pooled}(b) moves probability to or from the virtual novel class.

\paragraph{Regions without known classes (\cref{sec:rule}).} Fix $x$ with $p^*(\ell\mid x)=0$ for every known $\ell\preceq y$, so every image there is labelled $y$. The partial-label term $-\log Q_y$ depends on the nodes below $y$ only through their sum, so every split of $Q_y$ gives the same value. The coarsening term is $-\log(p_y+\sum_\ell w_\ell p_\ell)$ with $p_y+\sum_\ell p_\ell\le1$. The argument of the logarithm is linear in the probabilities, with coefficient $1$ for $p_y$ and $w_\ell<1$ for the leaves. Over the simplex it is therefore uniquely maximised at the vertex $p_y=1$. $\square$

\paragraph{Finite-sample identification (\cref{sec:finite}).} Arbor uses the virtual novel class only in superclasses that contain a class without fine-grained labels. Each of a known class's $N$ training images is labelled coarsely with probability $q$, independently, so the class receives no fine-grained label with probability $q^N$, and a superclass with known classes of sizes $N_1,\dots,N_K$ is assigned a virtual novel class with probability $1-\prod_k(1-q^{N_k})$. Computed from each class's actual training count and averaged over the superclasses with two or more classes and over three seeds, the predicted fraction matches the measured one at $q=0.9$ to within a point (\cref{tab:virtualfrac}).
\begin{table}[t]\centering\scriptsize\resizebox{\columnwidth}{!}{\begin{tabular}{lccccc}\toprule
 & CUB & Cars & CIFAR-100 & Aircraft & iNat-2019\\\midrule
median training images per class & $42$ & $59$ & $420$ & $70$ & $150$ (min $13$)\\
superclasses with a virtual novel class, predicted (\%) & $5.8$ & $0.8$ & $0.0$ & $0.2$ & $4.0$\\
superclasses with a virtual novel class, measured (\%) & $5.4$ & $0.0$ & $0.0$ & $0.0$ & $3.2$\\\bottomrule\end{tabular}}
\caption{\textbf{Fraction of superclasses assigned a virtual novel class} under random coarse labels ($q=0.9$, every class known), among those with two or more classes: predicted from each class's training count, and measured; three seeds.}\label{tab:virtualfrac}\end{table}
\paragraph{Power of a count test for novel classes.} Coarse-label counts reveal a novel class only in large classes. Under coarsening at random with probability $q$, a superclass's $N$ images split into $C\sim\mathrm{Bin}(N,q)$ coarse and $F=N-C$ fine; the excess of coarse labels over what the fine ones predict, $(C-qN)/(1-q)$, has mean zero and variance $Nq/(1-q)\approx qF/(1-q)^2$, and a novel class with $N'$ images shifts it by $\zeta\approx N'(1-q)/\sqrt{qF}$ standard deviations. From the training counts at $q=0.9$, the median shift over superclasses with a novel class is $1.6$ on CUB, $2.0$ on Cars, $2.4$ on Aircraft and $3.4$ on CIFAR-100, and with unit noise a test at $\zeta\ge3$ would detect $16$, $22$, $36$ and $66\%$ of them (seed $0$), even before correcting for testing every superclass. On the three fine-grained benchmarks a count test therefore misses most novel classes; CIFAR-100, with about $420$ images per class, is the one benchmark on which the test has power. The sample identifies the leaves that received no fine-grained label; this set is the plug-in estimate of $\mathcal U_y$, and the virtual novel class is placed on it.

\section{Lemmas and proofs}\label{app:lemmas}
\begin{figure}[t]\centering\includegraphics[width=0.85\columnwidth]{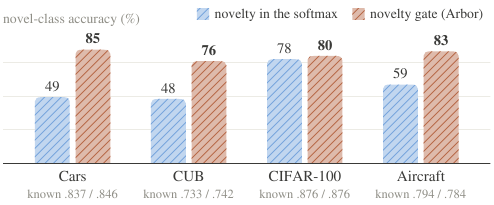}
\caption{\textbf{Virtual novel class as a softmax node and as a gate} (\cref{prop:decouple}). Novel-class accuracy in the main protocol with the virtual novel class as a node of the softmax and with its own gate (Arbor). Below each pair: known-class accuracy (softmax / gate). Three seeds; the rows ``Arbor'' and ``novelty in the softmax'' of \cref{tab:abl}.}\label{fig:decouple}\end{figure}
\paragraph{Known classes with few fine-grained labels.} A known class with few fine-grained labels can have its images assigned to a sibling class with more. Take two sibling classes $a$ and $b$ with features $x=\mu_c+\xi$, $\xi\sim\mathcal N(0,\varepsilon^2I)$. Let $W$ be a fixed projector with $T=\mathrm{tr}(WW^\top)$, and let $\Delta=\|W(\mu_b-\mu_a)\|^2$ be the projected margin. The prototypes $\phi_a,\phi_b$ equal the projected means of the $n_a$ and $n_b$ fine-labelled images. A coarse-labelled image of $b$ is assigned, within the superclass, to the nearer prototype.
\begin{lemma}[Misassignment between siblings with unequal label counts]\label{prop:absorb}
With $E_c=\|Wx-\phi_c\|^2$, $\ \mathbb E[E_a-E_b]=\Delta-\varepsilon^2T\,(1/n_b-1/n_a)$. With $n_a=n_b$, the expected margin is positive for every common count. With $n_a\to\infty$, the crossover is a threshold on $n_b$ alone. For $WW^\top=I_k$, the probability that a coarse-labelled image of $b$ is assigned to $a$ is approximately $\Phi\big(-(\gamma-k(1/n_b-1/n_a))/\sqrt v\big)$ with $\Phi$ the standard normal distribution function, $\gamma=\Delta/\varepsilon^2$ and $v=4\gamma(1+1/n_a)+4k(1/n_a+1/n_b)+2k(1/n_a^2+1/n_b^2)$.
\end{lemma}
The expected margin depends on $1/n_b-1/n_a$ rather than on $n_b$ alone, so a class with two fine-labelled images loses far fewer images to a sibling that also has two than to one that keeps all of its labels. \Cref{app:weak} tests the prediction.

\paragraph{Proof.} Write $\tilde\xi = W\xi$ and $h_c = W\bar\xi_c$ for the projected noise of the image and of the mean of class $c$'s fine-labelled images. They are independent, with covariances $\varepsilon^2WW^\top$ and $\tfrac{\varepsilon^2}{n_c}WW^\top$. Then $Wx - \phi_b = \tilde\xi - h_b$ and $Wx - \phi_a = Wd + \tilde\xi - h_a$ with $d = \mu_b - \mu_a$. Hence $\mathbb E[E_b] = \varepsilon^2(1+1/n_b)T$ and $\mathbb E[E_a] = \Delta + \varepsilon^2(1+1/n_a)T$. For $WW^\top = I_k$, expand $E_a - E_b = \Delta + 2\langle Wd, \tilde\xi - h_a\rangle + 2\langle \tilde\xi, h_b - h_a\rangle + \|h_a\|^2 - \|h_b\|^2$. The three random terms are pairwise uncorrelated, since each pairing is odd in $\tilde\xi$, $h_a$ or $h_b$. Their variances are $4\Delta\varepsilon^2(1+1/n_a)$, $4\varepsilon^4k(1/n_a+1/n_b)$ and $2\varepsilon^4k(1/n_a^2+1/n_b^2)$. A normal approximation to their sum gives the display; the exact law is a generalised chi-square, and the approximation is accurate for large $k$ or large $\gamma$. $\square$ \emph{Remark.} The lemma describes the mechanism in a frozen model, in which the projector is fixed and the biases ignored; in Arbor both train, and the fine-labelled images of the class with more fine-grained labels also train $W$, so the lemma does not bound the trained model.

\paragraph{The unused softmax term.} The term $r(y\mid x)$ of a gated superclass enters no label's likelihood, so the loss drives it to zero. Until then $P$ sums to $1-\sum_y r(y\mid x)$, the sum over the gated superclasses, a per-image constant that affects neither the most probable class nor the readout of \cref{eq:readout}, and reported probabilities are renormalised.

\paragraph{Proof of \cref{prop:decouple}.} The coarse-label term of \cref{eq:loss} splits through the identity $P(y\mid x)+\hat q\sum_{\ell\preceq y}P(\ell\mid x)=m_y\big(\hat q+(1-\hat q)\nu_y\big)$ into $-\log m_y$ and $-\log(\hat q+(1-\hat q)\nu_y)$, and the fine-grained term as $-\log r(\ell\mid x)-\log(1-\nu_y)$. Since \cref{eq:gate} can represent every node posterior within $y$, \cref{thm:pooled} applies to the gated model unchanged, with $\nu_y=\max\{0,(c-\hat q)/(1-\hat q)\}$ at the optimum.

\paragraph{Gradients of Arbor's loss.} \begin{lemma}[Gradients of Arbor's loss]\label{prop:em}
For a coarse label $y$, the gradient of Arbor's loss on the logit $o_n=-\|z-\phi_n\|^2/\tau+b_n$ of \cref{eq:softmax} is $r_n(1-\mathbf 1[n\preceq y]/m_y)$, and, in a superclass with a virtual novel class, on the gate's logit $\eta_y=g_y^\top z+\beta_y$ it is $-(1-\hat q)\nu_y(1-\nu_y)/(\hat q+(1-\hat q)\nu_y)$; for a fine-grained label under $y$, the gradient on $\eta_y$ is $\nu_y$. With the virtual novel class as a node of the softmax, the gradient on $o_n$ is $r_n(1-w_n/Q_w)$ with $Q_w=\sum_m w_mr_m$, and on an image whose mass lies under $y$ a descent step of size $\alpha$ changes the log-odds of two known siblings $\ell$ and $\ell'$ by $-\alpha(1-\hat q)\,(r_y/Q_w)\,(r_\ell-r_{\ell'})$.
\end{lemma}
A coarse-labelled image therefore raises the gate's logit and the logit of every class of its superclass, each by its probability, as in expectation--maximisation, and the gate never enters the softmax's gradient through the classifier head, although both share the embedding. A single softmax flattens the known classes of a superclass on every coarse-labelled image to which it gives novelty mass (row ``novelty in the softmax'' of \cref{tab:abl}; \cref{fig:decouple}). \paragraph{Proof.} With $r=\mathrm{softmax}(o)$ and $m_y=\sum_{\ell\preceq y}r_\ell$, $\partial(-\log m_y)/\partial o_n=r_n-\mathbf 1[n\preceq y]\,r_n/m_y$, and the gate's term $-\log(\hat q+(1-\hat q)\sigma(\eta_y))$ has derivative $-(1-\hat q)\nu_y(1-\nu_y)/(\hat q+(1-\hat q)\nu_y)$ in $\eta_y$; for a fine-grained label the gate's term is $-\log(1-\sigma(\eta_y))$, with derivative $\nu_y$. By \cref{prop:decouple} the two terms share no parameter of the head. For a single softmax, $\partial(-\log Q_w)/\partial o_n=r_n-w_nr_n/Q_w=r_n(1-w_n/Q_w)$. For two known siblings, $w_\ell=w_{\ell'}=\hat q$, so a descent step changes $o_\ell-o_{\ell'}$ by $-\alpha(1-\hat q/Q_w)(r_\ell-r_{\ell'})$. When the image's mass lies under $y$, $r_y+\sum_{\ell\preceq y}r_\ell=1$ and $Q_w=r_y+\hat q(1-r_y)$, so $1-\hat q/Q_w=(1-\hat q)\,r_y/Q_w$. $\square$

\paragraph{Tree prototypes against free prototypes.} Since $\|z\|^2$ cancels in the softmax, the distance logits of \cref{eq:softmax} are affine in $z$~\cite{snell2017prototypical}, as in a linear classifier, so the tree is an inductive bias, not extra capacity. The linear classifier of \cref{tab:abl} is trained with Arbor's loss and gate, its learning rate ($\times10$ or $\times100$) and embedding penalty ($\rho\in\{0.01,0.03\}$) chosen on validation. Class by class, the tree adds $0.2$--$2.6$ points of known-class accuracy over free prototypes in every third of the classes by number of fine-grained labels on Cars, CUB and Aircraft, and is within half a point of them on CIFAR-100.

\begin{lemma}[Unbounded growth of the embedding]\label{prop:radius}
Place the prototypes on the tree with edge length $L$, $\phi_\ell=\phi_{\mathrm{pa}(\ell)}+L\,u_\ell$ with $u_\ell$ the unit edge direction of leaf $\ell$, and let the leaf be extreme in that direction, $\langle u_\ell,\phi_n-\phi_\ell\rangle<0$ for every $n\ne\ell$. Then the cross-entropy of a leaf label under the softmax $r$ (\cref{eq:softmax}) decreases in the displacement $\delta$ of the embedding beyond the leaf along that direction, for every $\delta$; in a superclass with a novel class, the lemma does not cover the gate's term of the loss. With an L2 penalty $\rho\|z\|^2$, every stationary point along this direction has $\langle z,u_\ell\rangle\le M/(\rho\tau)$, that is $\delta\le\max\{0,M/(\rho\tau)-\langle \phi_\ell,u_\ell\rangle\}$, with $M=\max_{n\ne\ell}\langle u_\ell,\phi_\ell-\phi_n\rangle$.
\end{lemma}
Without the penalty the embedding grows: at $\rho=0$ in the mixed-granularity protocol, Arbor's mean embedding norm on Cars is $9.9$ against $1.9$ at its validation-selected penalty, its known-class accuracy is $0.8$--$4.4$ points lower on all four benchmarks, and it recognises $2$--$3$ points fewer novel classes on Cars, CUB and Aircraft. The penalty acts on the baselines in the other direction. With $\rho=0.03$, every class known and $90\%$ coarse labels, a flat softmax puts $95$--$98\%$ of a superclass's test predictions on its most-predicted leaf on Cars, CUB and Aircraft, against $53$--$67\%$ at $\rho=0$ and $52$--$69\%$ for Arbor at either value, so the penalty collapses linear logits but not Arbor's distances. Every classifier therefore chooses its $\rho$ on validation, Arbor and every other classifier with a virtual novel class from $\{0.01,0.03\}$ and every classification baseline from $\{0,0.01,0.03\}$, with the exceptions listed in \cref{app:data}. The exclusion of $\rho=0$ for the classifiers with a virtual novel class also rested on test results (\cref{app:data}).

\paragraph{Proof.} Write $\phi_\ell=\phi_{\mathrm{pa}(\ell)}+L\,u_\ell$ and $z=\phi_\ell+\delta\,u_\ell$ with $\delta\ge0$. Then $\|z-\phi_\ell\|^2 = \delta^2$, $\|z-\phi_{\mathrm{pa}(\ell)}\|^2 = (L+\delta)^2$, and $o_\ell - o_{\mathrm{pa}(\ell)} = (2L\delta + L^2)/\tau + b_\ell - b_{\mathrm{pa}(\ell)}$, which increases in $\delta$. The derivative of $-\log r_\ell$ with respect to $\delta$ is $\sum_{n\ne\ell}r_n\,\partial(o_n - o_\ell)/\partial\delta = (2/\tau)\sum_{n\ne\ell}r_n\langle u_\ell, \phi_n - \phi_\ell\rangle$. For the parent term it equals $-r_{\mathrm{pa}(\ell)}\,2L/\tau$. For every other node it is negative if and only if $\langle u_\ell, \phi_n - \phi_\ell\rangle < 0$, the extreme-point condition in the statement. The condition holds for the parent and for the siblings, which sit at the same distance from the same parent; it can fail for other ancestors, or for a prototype of another superclass that lies beyond $\phi_\ell$ along $u_\ell$, and without it the loss can increase in $\delta$ through that node. Under the condition the derivative is negative for every $\delta$, so without a penalty on $\|z\|$ the loss decreases monotonically in $\delta$. The biases alone cannot stop the growth, because the gradient on $b_{\mathrm{pa}(\ell)}$ vanishes as $r_{\mathrm{pa}(\ell)}\to0$ at the same rate. The penalty is measured from the origin of the embedding space, so $d(\rho\|z\|^2)/d\delta=2\rho\langle z,u_\ell\rangle=2\rho(\langle \phi_\ell,u_\ell\rangle+\delta)$. A stationary $\delta>0$ therefore satisfies $2\rho(\langle \phi_\ell,u_\ell\rangle+\delta) = (2/\tau)\sum_{n\ne\ell}r_n\langle u_\ell,\phi_\ell-\phi_n\rangle \le (2/\tau)\,M\sum_{n\ne\ell}r_n\le 2M/\tau$, hence $\delta\le M/(\rho\tau)-\langle \phi_\ell,u_\ell\rangle=M/(\rho\tau)-L-\langle \phi_{\mathrm{pa}(\ell)},u_\ell\rangle$. The bound is $M/(\rho\tau)-L$ when the parent's prototype is orthogonal to the leaf's edge, and larger when the edge points back towards the origin. $\square$ \emph{Remark.} The argument restricts $z$ to the ray $\phi_\ell + \delta\,u_\ell$ with prototypes and biases fixed; it bounds the growth of the embedding along the edge and does not characterise the unconstrained optimum.

\paragraph{The uniform target.} A leaf model can be given a virtual novel class indirectly: coarse labels are trained against the uniform distribution over their descendants, and the superclass is read from the marginal. This is the averaging baseline of partial-label learning~\cite{cour2011partial}.
\begin{lemma}[Cross-entropy against a uniform target]\label{prop:maxent}
Cross-entropy against the uniform target over the $K$ leaves under $y$ equals $-\log Q_y+\mathrm{KL}(\mathrm{Unif}\,\|\,p(\cdot\mid y,x))+\log K$, which is minimised at equal child probabilities. On the coarse-labelled images of a class $b$ that also has fine-labelled images, the same term biases $p(b\mid y,x)$ towards $1/K$, and the optimal margin between $b$ and its siblings goes to zero as $q\to1$.
\end{lemma}
\paragraph{Proof.} Cross-entropy against the uniform target $\mathrm{Unif}$ over the $K$ leaves under $y$ is $-\frac1K\sum_{\ell\preceq y}\log p_\ell = -\log Q_y - \frac1K\sum_{\ell\preceq y}\log p(\ell\mid y,x) = -\log Q_y + \mathrm{KL}(\mathrm{Unif}\,\|\,p(\cdot\mid y,x)) + \log K$ with $p(\ell\mid y,x) = p_\ell/Q_y$. The KL term is minimised at $p(\cdot\mid y,x) = \mathrm{Unif}$. Take a class $b$ with a fraction $q$ of its images labelled with the superclass. The loss summed over $b$'s images contains $(1-q)\,\mathbb E[-\log p_b] + q\,\mathbb E[\mathrm{KL}(\mathrm{Unif}\,\|\,p(\cdot\mid y,x))]$. The first term decreases in the margin $o_b - o_a$ and the second increases. The optimal margin is therefore finite and goes to zero as $q\to1$. $\square$

\section{Datasets and model selection}\label{app:data}
\paragraph{Metrics.} With $\mathrm{anc}(\ell)$ the class and its ancestors, the exact-class rate and the incorrect rate of \cref{sec:setup} are $\mathrm{ex}=\tfrac{1}{|\mathcal D_{\mathrm K}|}\sum_{x\in\mathcal D_{\mathrm K}}\mathbf 1[\hat n(x)=\ell(x)]$ and $\mathrm{inc}=\tfrac{1}{|\mathcal D_{\mathrm K}\cup\mathcal D_{\mathrm U}|}\sum_{x}\mathbf 1[\hat n(x)\notin\mathrm{anc}(\ell(x))]$. The leaf column of \cref{tab:merged_full} is the share of $x\in\mathcal D_{\mathrm U}$ with $\mathrm{depth}(\hat n(x))=D$.

\paragraph{Splits and sizes.} The benchmarks are split $70/10/20$ into training, validation and test; the embedding has $k=128$ dimensions ($64$ on CIFAR-100); iNat-2019 has $1{,}010$ species in $72$ genera and $34$ orders.

Stanford Cars has no official hierarchy. We parse each class name into body type and make, giving body type $\to$ (body type, make) $\to$ model. These levels have 10, 110 and 196 nodes, and 64 of the 110 superclasses have a single model. CUB-200-2011 is grouped by the order and family that each species' common-name suffix implies: 13 orders, and 70 groups at the superclass level, 33 of them with a single species. These groups collect species that share a common-name suffix; they are not taxonomic families. CIFAR-100's 20 superclasses are grouped into four (animals; plants and nature; household; vehicles and structures). Every superclass has five classes. FGVC-Aircraft uses its official three levels: 30 manufacturers and 70 families (the dataset's own name for this level), 54 of them with a single variant. In every protocol with one novel class per superclass, one leaf in every superclass with two or more leaves is a novel class. These are 46 of 196 classes on Cars, 37 of 200 on CUB, 20 of 100 on CIFAR-100 and 16 of 100 on Aircraft. All trees and index files are released.

\paragraph{Novel classes and scoring.} In the mixed-granularity protocols the novel class of each superclass is its first listed leaf, the same for every seed, except in the protocol that draws it at random per seed (\cref{app:moreprotocols}); the seeds change the splits and which images are coarse. A prediction of a novel class's own leaf is not counted as incorrect, but it does not recognise the novel class and counts in the leaf column; no classifier receives a fine-grained label for it. The taxonomy lists the novel classes, so every classifier's leaf softmax includes them. With mixed-granularity labels, the flat softmax sends at most $0.6\%$ of known-class test images to a novel leaf, the uniform target at most $4.0\%$ and Arbor at most $0.2\%$. Restricting the argmax to known leaves moves Arbor's known-class gap to the most accurate classification baseline by at most $0.2$ points; with that restriction the uniform target reaches $.797$ on Aircraft, above Arbor's $.784$.

\begin{table*}[t]\centering\scriptsize\setlength{\tabcolsep}{2.4pt}
\begin{tabular}{lccccccccccccccc}
\toprule
 & \multicolumn{3}{c}{Cars} & \multicolumn{3}{c}{CUB} & \multicolumn{3}{c}{CIFAR-100} & \multicolumn{3}{c}{Aircraft} & \multicolumn{3}{c}{iNat-2019} \\
\cmidrule(lr){2-4}\cmidrule(lr){5-7}\cmidrule(lr){8-10}\cmidrule(lr){11-13}\cmidrule(lr){14-16}
method & novel$\uparrow$ & leaf$\downarrow$ & known$\uparrow$ & novel$\uparrow$ & leaf$\downarrow$ & known$\uparrow$ & novel$\uparrow$ & leaf$\downarrow$ & known$\uparrow$ & novel$\uparrow$ & leaf$\downarrow$ & known$\uparrow$ & novel$\uparrow$ & leaf$\downarrow$ & known$\uparrow$ \\
\midrule
Flat softmax + HSC~\cite{goren2024hsc} & 6 & 83 & .844 & 12 & 85 & .715 & 17 & 75 & .888 & 7 & 82 & \textbf{.791} & 34 & 61 & .639 \\
Conditional softmax~\cite{redmon2017yolo9000,valmadre2022hierarchical} & 6 & 83 & .814 & 14 & 83 & .745 & \textbf{81} & \textbf{12} & .876 & 8 & 77 & .768 & 29 & 65 & .622 \\
Per-node sigmoids~\cite{valmadre2022hierarchical} & 14 & 70 & .826 & 17 & 82 & .757 & 56 & 37 & .893 & 17 & 74 & .762 & 28 & 70 & .598 \\
Per-level + fallback~\cite{park2025hcast,garg2022haf,chu2026hieruq} & 6 & 81 & .839 & 10 & 86 & \textbf{.759} & 34 & 56 & \textbf{.908} & 6 & 83 & .780 & 30 & 65 & \textbf{.646} \\
Hierarchical CE~\cite{bertinetto2020making} & 10 & 80 & .804 & 14 & 83 & .707 & 36 & 56 & .873 & 11 & 80 & .762 & 30 & 66 & .607 \\
Soft labels~\cite{bertinetto2020making} & 6 & 84 & \textbf{.851} & 12 & 83 & .724 & 32 & 62 & .884 & 8 & 82 & .784 & 29 & 66 & .637 \\
HypStructure~\cite{zeng2024hypstructure} & 8 & 80 & .834 & 11 & 88 & .713 & 23 & 70 & .887 & 7 & 82 & .767 & 35 & 61 & .616 \\
\midrule
Uniform target & 39 & 57 & .799 & 24 & 76 & .637 & 67 & 30 & .753 & 42 & 52 & .784 & 46 & 52 & .467 \\
Hierarchical novelty detection~\cite{lee2018hierarchical} & 68 & 31 & .738 & 45 & 53 & .637 & 78 & 19 & .816 & 57 & 39 & .708 & 43 & 55 & .441 \\
\midrule
\rowcolor{ourspastel} \textbf{Arbor (ours)} & \textbf{85} & \textbf{13} & .846 & \textbf{76} & \textbf{23} & .742 & 80 & 16 & .876 & \textbf{83} & \textbf{10} & .784 & \textbf{68} & \textbf{31} & .594 \\
\bottomrule
\end{tabular}

\caption{\textbf{\Cref{tab:merged} with the leaf column.} Leaf: novel-class images assigned to a leaf (\%), including the novel class's own leaf; on Cars, CUB and Aircraft Arbor sends $10$--$23\%$ of novel-class images to a leaf against $70$--$88\%$ for the classification baselines. Other columns and the bold rule as in \cref{tab:merged}.}
\label{tab:merged_full}\end{table*}
\paragraph{Baselines.} \Cref{tab:baselines_mech} lists the baselines by group and says how each uses the taxonomy. Among the hierarchical classifiers of the literature, hierarchical cross-entropy and soft labels~\cite{bertinetto2020making} weight leaf errors by tree distance, Deep-RTC~\cite{wu2020deeprtc} trains node-conditional softmaxes and rejects top-down, HAF~\cite{garg2022haf}, BiLT~\cite{lu2025bilt} and H-CAST~\cite{park2025hcast} add one classifier per level, HierUQ~\cite{chu2026hieruq} adds a confidence-aware fallback, and hyperbolic~\cite{sarkar2011tree,nickel2017poincare,ganea2018cones,khrulkov2020hyperbolic} and prototype~\cite{snell2017prototypical,mettes2019hyperspherical,zeng2024hypstructure} embeddings put the hierarchy in the representation. Our per-level + fallback baseline reimplements the per-level classifiers of HAF, H-CAST and HierUQ on the shared backbone; \cref{app:official} runs released implementations. HypStructure represents the hyperbolic and hierarchy-aware prototype embeddings of~\cite{dhall2020hierarchical,desai2023meru,pal2025hycoclip,atigh2021busemann,barz2019hierarchy,garnot2021metric}; \cref{app:pu} defines the Elkan--Noto correction and the two-stage baseline.
\begin{table*}[t]\centering\scriptsize\setlength{\tabcolsep}{3pt}
\begin{tabular}[t]{@{}l>{\raggedright\arraybackslash}p{4.3cm}@{}}\toprule
baseline & how it uses the taxonomy\\\midrule
\multicolumn{2}{@{}l}{\emph{classification}}\\
flat softmax + HSC~\cite{goren2024hsc} & flat softmax; thresholded node probabilities\\
hierarchical CE~\cite{bertinetto2020making} & loss weighted along the path\\
soft labels~\cite{bertinetto2020making} & targets softened by tree distance\\
conditional softmax~\cite{redmon2017yolo9000,valmadre2022hierarchical} & one softmax per internal node\\
per-node sigmoids~\cite{valmadre2022hierarchical} & one sigmoid per node\\
per-level + fallback~\cite{chu2026hieruq,park2025hcast,garg2022haf} & one classifier per level; falls back below a confidence threshold\\
HypStructure~\cite{zeng2024hypstructure} & tree-structured hyperbolic regulariser\\
\bottomrule
\end{tabular}\hfill
\begin{tabular}[t]{@{}l>{\raggedright\arraybackslash}p{4.3cm}@{}}\toprule
baseline & how it uses the taxonomy\\\midrule
\multicolumn{2}{@{}l}{\emph{classification, special cases only}}\\
ensemble ($5\times$)~\cite{lakshminarayanan2017ensembles} & five flat softmax classifiers, averaged\\
hyperbolic Busemann$^\dagger$~\cite{atigh2021busemann} & hyperbolic prototypes\\
flat softmax over all nodes, min/max weighting & one weight per superclass for its coarse labels (\cref{tab:special_full})\\\midrule
\multicolumn{2}{@{}l}{\emph{novelty}}\\
uniform target (\cref{prop:maxent}) & coarse label: uniform over its classes\\
hierarchical novelty detection~\cite{lee2018hierarchical} & virtual novel class in the node softmax\\\midrule
\multicolumn{2}{@{}l}{\emph{post hoc}}\\
Elkan--Noto correction~\cite{elkan2008pu} & hierarchical novelty detection, corrected after training\\
two-stage & a classifier's class, or its superclass if the corrected novelty mass is high\\\bottomrule
\end{tabular}
\caption{\textbf{Baselines: how each uses the taxonomy.} HSC: hierarchical selective classification. $^\dagger$Also in the unseen-class protocol. Hierarchical novelty detection runs on Arbor's node classifier with every coarse label on the virtual novel class; the two-stage baseline uses the classification baseline with the highest validation accuracy (\cref{app:pu}).}\label{tab:baselines_mech}\end{table*}

\paragraph{Model selection.} Every hyperparameter is chosen on the validation split of the protocol reported. Validation images are coarsened like training images, so model selection and the operating point rest mainly on the fine-labelled validation images: at $90\%$ coarse labels, $129$, $97$, $481$ and $76$ on Cars, CUB, CIFAR-100 and Aircraft ($0.6$--$0.9$ per known class on the fine-grained benchmarks) and $2{,}506$ on iNat-2019. The hyperparameters are $\alpha_{\mathrm{HCE}}$ for hierarchical cross-entropy, $\beta_{\mathrm{soft}}$ for soft labels, the curvature of the hyperbolic classifier, the classifier learning rate, the embedding penalty $\rho\in\{0,0.01,0.03\}$ of every classification baseline and $\rho\in\{0.01,0.03\}$ of every classifier with a virtual novel class, and the variant of the flat softmax over all nodes; \cref{tab:selected} lists Arbor's selected $\rho$. The criterion is fine-grained accuracy on fine-labelled validation images. In the protocol of novel classes with every other label fine-grained, this criterion does not depend on novel-class images, so we add the superclass rate on coarse-labelled validation images, the fraction assigned to their superclass. The grids have these exceptions:
\begin{itemize}[leftmargin=1.2em,itemsep=1pt,topsep=2pt]
\item on the merge, hierarchical CE ($\alpha_{\mathrm{HCE}}=0.4$), soft labels ($\beta_{\mathrm{soft}}=30$) and HypStructure ran at $\rho=0$ only;
\item on iNat-2019, novelty detection ran at $\rho=0.03$ only, and hierarchical CE, soft labels and HypStructure without $\rho=0.03$; the diagnostic rows of \cref{tab:inat} use seed $0$ and $\rho=0.01$;
\item with unseen classes, the baselines ran at $\rho=0.03$ only;
\item on naturally coarse labels, soft labels ran at $\rho=0$ only, HypStructure without $\rho=0.03$, novelty detection at $\rho=0.03$ only, and the linear classifier with Arbor's loss and gate in one configuration ($\times10$, $\rho=0.01$); the classification baselines also ran with the $\times10$ learning rate, which was never selected on validation;
\item the $\times10$ learning rate was not crossed with every variant of hierarchical CE, soft labels and HypStructure;
\item the baselines of the iNat-2019 special cases and of the unseen-class protocol use the $\times1$ learning rate only.
\end{itemize}
Within these grids no test label is used; the choices that shaped them are listed below.

\paragraph{Choosing the novelty gate.} Fine-grained validation accuracy does not depend on novel-class images, so the novelty gate is chosen with pseudo-novel classes. In each superclass with at least three classes, one more known class (a different random class per seed) keeps only its superclass label in training and validation: Cars $23$ classes and $193$ validation images per seed, CUB $21$ and $123$, CIFAR-100 $20$ and $1{,}200$, Aircraft $6$ and $60$. Each candidate is scored by the mean of fine-grained validation accuracy and the share of these classes' validation images predicted as their superclass, at the operating point. The gate scores $84.6$ against at most $78.4$ for any single-softmax weighting and is chosen on all four benchmarks (\cref{tab:pseudonovel}); fine-grained validation accuracy alone is up to $2.8$ points lower for the gate. Fine-grained validation accuracy alone would favour the softmax-node variant of \cref{tab:abl}, whose validation accuracy is higher on three of the four benchmarks and whose novel-class accuracy is lower on all four (\cref{tab:pseudonovel}).
\begin{table}[t]\centering\scriptsize\setlength{\tabcolsep}{3pt}\resizebox{\columnwidth}{!}{\begin{tabular}{lccccccccc}
\toprule
 & \multicolumn{2}{c}{Cars} & \multicolumn{2}{c}{CUB} & \multicolumn{2}{c}{C100} & \multicolumn{2}{c}{Aircraft} & \\
\cmidrule(lr){2-3}\cmidrule(lr){4-5}\cmidrule(lr){6-7}\cmidrule(lr){8-9}
variant & F & PN & F & PN & F & PN & F & PN & score \\
\midrule
\rowcolor{ourspastel} \textbf{Arbor} (novelty gate, pooled $\hat q$) & .890 & \textbf{87} & .732 & \textbf{83} & .890 & \textbf{89} & .815 & \textbf{84} & \textbf{84.6} \\
virtual novel class in the softmax, pooled $\hat q$ & .889 & 34 & .764 & 48 & .895 & 76 & .834 & 77 & 71.6 \\
virtual novel class in the softmax, $\hat q$ on untrained leaves & .890 & 71 & .756 & 74 & .897 & 76 & .815 & 39 & 74.4 \\
pooled $\hat q$ without the restriction to gated superclasses & \textbf{.892} & 30 & .752 & 64 & .895 & 76 & .809 & 40 & 68.1 \\
$\hat q$ pooled per superclass & .877 & 82 & \textbf{.768} & 43 & .894 & 75 & .825 & 53 & 73.7 \\
per-class $\hat q_\ell$ & .883 & 75 & .716 & 48 & .886 & 69 & \textbf{.849} & 50 & 71.9 \\
share-weighted & .839 & 61 & .753 & 48 & \textbf{.911} & 1 & .825 & 61 & 62.8 \\
min/max weighting, $\kappa=10$ & .831 & 79 & .673 & 63 & .846 & 83 & .802 & 83 & 77.9 \\
min/max weighting, $\kappa=0$ & .816 & 77 & .681 & 70 & .846 & 83 & .821 & 81 & 78.4 \\
\bottomrule
\end{tabular}
}
\caption{\textbf{The three choices first made on test results, redone on validation with pseudo-novel classes} (one known class per superclass with at least three classes, coarse-only in training and validation). Rows: the gate and every candidate coarse-label weighting as a single softmax, each at the $\rho\in\{0,0.01,0.03\}$ with the highest fine-grained validation accuracy. F: fine-grained validation accuracy; PN: pseudo-novel validation images predicted as their superclass at the operating point (\%); score: mean of F and PN over the four benchmarks; three seeds.}\label{tab:pseudonovel}\end{table}

\paragraph{Choices that used test results, redone on validation.} Three choices were first made on test results. (i) Arbor's coarse-label weights were chosen among candidate weightings (min/max with and without a discount $\kappa$, share-weighted, per-class, pooled per superclass, and pooled over the dataset with and without the restriction to superclasses with a class without fine-grained labels) by known- and novel-class accuracy on the test images of the main and special-case protocols. (ii) The novelty gate was first preferred on test results. (iii) The embedding-penalty grid $\{0.01,0.03\}$ of classifiers with a virtual novel class excludes $\rho=0$ because, under one of the candidate weightings, $\rho=0$ lowered novel-class accuracy on test images. All three were then redone with the validation rule above and no test label (\cref{tab:pseudonovel}): every candidate weighting as a single softmax, the gate, and $\rho\in\{0,0.01,0.03\}$. The rule selects the gate on every benchmark (score $84.6$ against $78.4$ for the best single-softmax weighting) and, with $\rho=0$ allowed, selects $\rho=0.01$ or $0.03$ for the gate on every benchmark. Among the single-softmax weightings alone it ranks the min/max weighting first ($78.4$) and the pooled-with-restriction weighting sixth ($71.6$), so that choice, taken on its own, is not what validation would have made. With $\rho=0$ allowed, fine-grained validation accuracy selects $\rho=0$ for several single-softmax variants, where it lowers their pseudo-novel score (the effect of \cref{prop:radius}); for the gate it never does. Thresholds matched to another method's exact-class rate are set on validation, except along the swept curves (\cref{fig:curves,fig:tradeoff}) and in the matched comparison on the merge (\cref{app:merge}), which read the test curves. The gate's adoption on the six further settings ($50\%$ coarse labels, two novel classes, class-dependent coarsening, a random novel class, the merge and iNat-2019) followed a criterion fixed before those runs: known-class accuracy and novel-class accuracy at equal exact-class rate. The naturally coarse iNaturalist labels, the iNat-2019 special cases, the unseen-class protocol, the protocol of \cref{app:weak} and the CUB official split were not used for any of the three choices. Their results are in \cref{app:natural,app:full,app:unseen,app:weak,app:official}.
\begin{table}[t]\centering\scriptsize\setlength{\tabcolsep}{4pt}
\begin{tabular}{lcccc}
\toprule
protocol & Cars & CUB & C100 & Air. \\
\midrule
$90\%$ coarse, one novel class & .03 & .01 & .01 & .01 \\
$90\%$ coarse, two novel classes & .03 & .03 & .01 & .03 \\
$50\%$ coarse, one novel class & .03 & .03 & .03 & .03 \\
\midrule
all labels fine & .03 & .03 & .03 & .03 \\
$90\%$ coarse, no novel class & .01 & .01 & .01 & .03 \\
only novel classes coarse & .01 & .01 & .03 & .01 \\
\bottomrule
\end{tabular}
\caption{\textbf{Arbor's validation-selected embedding penalty} $\rho$, per protocol and benchmark. Top: mixed-granularity protocols (\cref{tab:merged,tab:twonever,tab:q05}); bottom: the special cases (\cref{tab:main}).}\label{tab:selected}\end{table}

\section{Predictions at the operating point}\label{app:op}
\begin{figure*}[t]\centering\includegraphics[width=\textwidth]{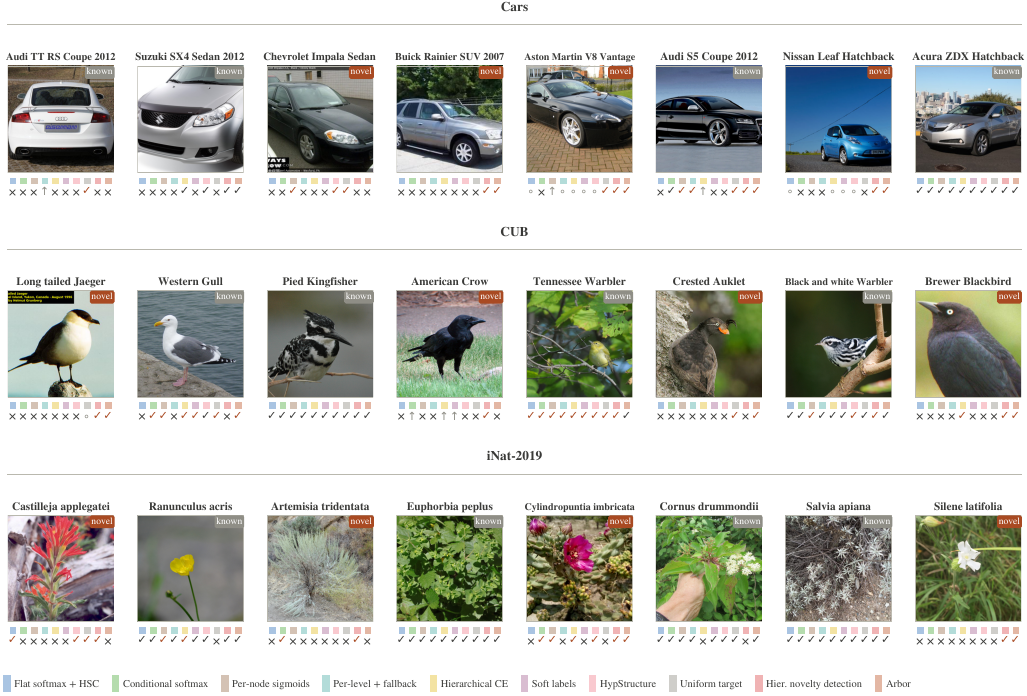}
\caption{\textbf{Test images drawn at random}, the three-benchmark version of \cref{fig:examples} (fixed seed, distinct classes; main protocol; eight per benchmark, half of novel and half of known classes). Under each image, one column per method in the legend's order, Arbor last: the colour names the method and the mark its answer at the shared operating point. Black $\checkmark$: the exact class. Orange $\checkmark$: the superclass, correct for a novel class and coarse but not wrong for a known one. $\times$: a wrong class. $\circ$: a novel class's own leaf, not recognised as novel. $\uparrow$: an ancestor above the superclass.}
\label{fig:examples_full}\end{figure*}
\begin{figure*}[t]\centering\includegraphics[width=0.84\textwidth]{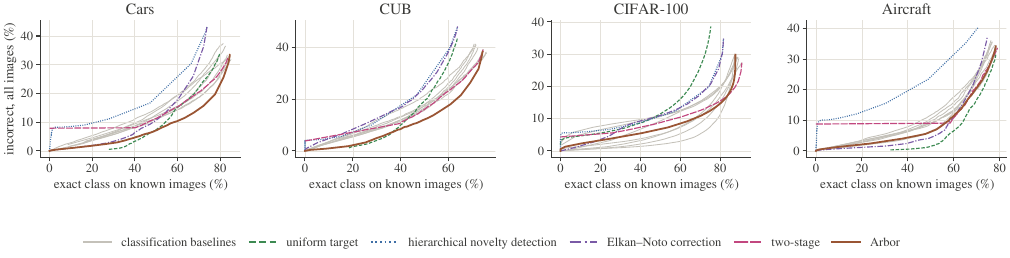}
\caption{\textbf{Incorrect predictions across every operating point}, mixed-granularity labels, three seeds: incorrect predictions over all test images against the exact-class rate on known-class images.}
\label{fig:tradeoff}\end{figure*}

With $\mathcal D_{\mathrm K}$ and $\mathcal D_{\mathrm U}$ the known- and novel-class test images, $\ell(x)$ the class of $x$, $\mathrm{pa}(\ell)$ its superclass and $\hat n(x)$ the predicted node, the two headline metrics are
{\small\begin{align*}
\mathrm{acc}_{\mathrm K}&=\tfrac{1}{|\mathcal D_{\mathrm K}|}\textstyle\sum_{x\in\mathcal D_{\mathrm K}}\mathbf 1\big[\arg\max_{\ell'}P(\ell'\mid x)=\ell(x)\big],\\
\mathrm{acc}_{\mathrm U}&=\tfrac{1}{|\mathcal D_{\mathrm U}|}\textstyle\sum_{x\in\mathcal D_{\mathrm U}}\mathbf 1\big[\hat n(x)=\mathrm{pa}(\ell(x))\big].
\end{align*}}
\Cref{tab:merged} reports known-class accuracy from the leaf posterior and novel-class accuracy from the node prediction at the same operating point, each method's threshold set on validation to a mean specificity of $0.9$. \Cref{fig:tradeoff} gives incorrect predictions across every operating point; in \cref{fig:curves}, Arbor's novel-class accuracy curve is the highest from $46\%$ exact-class rate upward on Cars, over $29$--$74\%$ on CUB, $60$--$78\%$ on Aircraft, and $41$--$50\%$ and $79$--$87\%$ on CIFAR-100. Reading known-class accuracy from the tree softmax $r$ instead of $P$ changes it by at most $0.8$ points (Cars; $0.3$ on Aircraft, $\le0.1$ elsewhere). \Cref{tab:op} scores the node prediction itself on known-class images at that point.

Fixing the mean specificity fixes the average depth of the predictions; it does not fix how many known-class images are predicted as their superclass, nor which ones. The baselines predict the superclass for their least confident images in every superclass. Arbor predicts the superclass for known-class images in the superclasses that contain a novel class, where its virtual novel class is used, and keeps leaf predictions in the other superclasses: its exact-class rate is $8$--$37$ points lower in superclasses with a novel class and $2$--$9$ points higher in the rest. On Cars, CUB and Aircraft, its rate of incorrect predictions on known-class images is at most $4.3$ points above the best classification baseline's, since the fine-grained predictions that fall back are mostly assigned to the true superclass; on CIFAR-100 this rate is $5.2$ points higher. Over all test images, Arbor makes incorrect predictions $7$--$12$ points less often on Cars, CUB and Aircraft, and $4.4$ points more often on CIFAR-100. At the exact-class rate of the classification baseline with the most novel classes, Arbor still recognises $68$--$70\%$ of novel-class images on Cars, CUB and Aircraft against $14$--$17\%$, and $76\%$ on CIFAR-100 against the conditional softmax's $81\%$ (\cref{tab:deployed}). Over three seeds, Arbor's margin in novel-class accuracy holds in every seed on Cars, CUB and Aircraft. The paired difference in known-class accuracy is $-0.5\pm1.1$ points on Cars, $-1.6\pm1.2$ on CUB, $-3.3\pm0.3$ on CIFAR-100 and $-0.7\pm0.7$ on Aircraft (mean $\pm$ population std over seeds).

\paragraph{What a superclass prediction contains.} A superclass prediction covers novel-class images and known-class images whose class is uncertain. In superclasses with a novel class, Arbor predicts the superclass for $22$--$71\%$ of known-class images, and novel-class images make up $38$--$48\%$ of these predictions against a base rate of $20$--$35\%$. The probability of its virtual novel class separates novel- from known-class images with an AUROC of $0.63$--$0.80$, against $0.47$--$0.72$ for the novelty-detection classifier's superclass probability. These predictions are never incorrect, since the superclass is an ancestor of the true class; their cost is the exact-class rate in \cref{tab:op}.

\begin{table*}[t]\centering\scriptsize\setlength{\tabcolsep}{4pt}\resizebox{0.74\textwidth}{!}{\begin{tabular}{lcccc}
\toprule
 & Cars & CUB & CIFAR-100 & Aircraft \\
\midrule
\multicolumn{5}{l}{\textit{one novel class per superclass (\cref{tab:merged})}: Arbor / best classification baseline} \\
exact class, known-class images & 60.5 / 77.0 & 58.1 / 62.7 & 68.3 / 78.4 & 61.7 / 72.8 \\
\quad in superclasses with a novel class & 37.4 / 72.7 & 48.7 / 56.3 & 68.3 / 78.4 & 14.1 / 51.5 \\
\quad in the other superclasses & 92.3 / 83.4 & 95.0 / 92.7 & none & 88.2 / 86.4 \\
incorrect, known-class images & 9.4 / 7.2 & 14.5 / 10.2 & 8.3 / 3.2 & 11.9 / 11.5 \\
incorrect, all images & 10.7 / 22.4 & 16.2 / 23.3 & 10.4 / 6.0 & 12.3 / 23.0 \\
\cmidrule(lr){1-5}
exact class, Arbor at the best baseline's exact-class rate & 76.5 / 77.0 & 67.4 / 62.7 & 78.6 / 78.4 & 74.4 / 72.8 \\
novel-class acc., Arbor at the best baseline's exact-class rate & 53.8 / 13.9 & 56.5 / 16.8 & 70.1 / 80.8 & 36.9 / 17.1 \\
\multicolumn{5}{l}{paired over three seeds, mean $\pm$ std (minimum), points:} \\
known-class acc., Arbor $-$ best (min) & $-0.5\pm1.1$ ($-1.7$) & $-1.6\pm1.2$ ($-2.5$) & $-3.3\pm0.3$ ($-3.6$) & $-0.7\pm0.7$ ($-1.6$) \\
novel-class acc., Arbor $-$ best (min) & $+71\pm2$ ($+69$) & $+57\pm4$ ($+52$) & $-1\pm1$ ($-2$) & $+66\pm2$ ($+65$) \\
\midrule
\multicolumn{5}{l}{\textit{two novel classes per superclass (\cref{tab:twonever})}: Arbor / best classification baseline} \\
exact class, known-class images & 66.1 / 74.6 & 55.6 / 62.9 & 66.6 / 82.4 & 67.5 / 69.1 \\
\quad in superclasses with a novel class & 12.2 / 69.7 & 36.4 / 52.9 & 66.6 / 82.4 & 7.7 / 44.0 \\
\quad in the other superclasses & 85.9 / 76.8 & 83.3 / 81.4 & none & 78.8 / 74.4 \\
incorrect, known-class images & 10.6 / 8.8 & 13.3 / 8.5 & 8.3 / 3.3 & 14.1 / 12.5 \\
incorrect, all images & 9.5 / 23.6 & 13.0 / 21.3 & 10.2 / 10.5 & 13.1 / 19.0 \\
\cmidrule(lr){1-5}
exact class, Arbor at the best baseline's exact-class rate & 75.7 / 74.6 & 63.0 / 62.9 & 81.7 / 82.4 & 75.2 / 69.1 \\
novel-class acc., Arbor at the best baseline's exact-class rate & 81.2 / 13.9 & 76.0 / 26.9 & 70.3 / 74.9 & 56.1 / 29.6 \\
\multicolumn{5}{l}{paired over three seeds, mean $\pm$ std (minimum), points:} \\
known-class acc., Arbor $-$ best (min) & $+0.9\pm0.4$ ($+0.3$) & $-3.7\pm2.0$ ($-6.5$) & $-3.1\pm0.2$ ($-3.4$) & $-0.7\pm0.6$ ($-1.2$) \\
novel-class acc., Arbor $-$ best (min) & $+77\pm3$ ($+73$) & $+57\pm2$ ($+56$) & $+12\pm2$ ($+10$) & $+59\pm11$ ($+44$) \\
\bottomrule
\end{tabular}
}
\caption{\textbf{Predictions at the operating point in the mixed-granularity protocols} (\%; DINOv2, three seeds). Each cell: Arbor / the classification baseline with the best value of that row (lowest for incorrect predictions). Paired rows compare known-class accuracy with the most accurate classification baseline, and novel-class accuracy with the best one in each seed. Rows at the same operating point use a mean specificity of $0.9$ on fine-labelled validation images. Rows at equal exact-class rate set Arbor's threshold on validation to the most exact classification baseline's; their right-hand values are the best of that row among the classification baselines at their own operating points. \Cref{tab:deployed} adds the novelty baselines. An incorrect prediction is a node that is neither the true class nor one of its ancestors.}\label{tab:op}\end{table*}

\section{The Elkan--Noto correction and the two-stage baseline}\label{app:pu}
The Elkan--Noto correction and the two-stage baseline of \cref{tab:pu} are defined here; \cref{tab:pu_full} is the full version of \cref{tab:pu}, and the configurations of \cref{fig:gains} follow.

\begin{table}[t]\centering\scriptsize\setlength{\tabcolsep}{3pt}
\resizebox{\columnwidth}{!}{\begin{tabular}{lcccc}
\toprule
 & Cars & CUB & C100 & Air. \\
\midrule
\multicolumn{5}{l}{\textit{Elkan--Noto correction / Arbor, each at its own operating point}} \\
known-class acc. & .738 / .846 & .637 / .742 & .816 / .876 & .708 / .784 \\
novel-class acc. & 77 / 85 & 68 / 76 & 78 / 80 & 78 / 83 \\
exact-class rate & 57 / 61 & 48 / 58 & 61 / 68 & 62 / 62 \\
\midrule
\multicolumn{5}{l}{\textit{Elkan--Noto correction / Arbor at the correction's exact-class rate}} \\
novel-class acc. & 77 / 87 & 68 / 86 & 78 / 82 & 78 / 83 \\
incorrect, all images & 15.4 / 9.6 & 22.5 / 12.1 & 13.8 / 9.4 & 13.3 / 12.2 \\
\midrule
\multicolumn{5}{l}{\textit{two-stage / Arbor, each at its own operating point}} \\
known-class acc. & .839 / .846 & .745 / .742 & .908 / .876 & .791 / .784 \\
novel-class acc. & 66 / 85 & 72 / 76 & 75 / 80 & 76 / 83 \\
incorrect, all images & 15.1 / 10.7 & 18.6 / 16.2 & 11.0 / 10.4 & 12.7 / 12.3 \\
\bottomrule
\end{tabular}}
\caption{\textbf{The Elkan--Noto correction and the two-stage baseline} against Arbor, full version of \cref{tab:pu}; main protocol, three seeds (known-class accuracy as a fraction, other rows in \%). The correction is the novelty-detection classifier corrected after training (Elkan--Noto); the two-stage baseline takes its known classes from the classification baseline with the highest validation accuracy and its novelty from the correction.}\label{tab:pu_full}\end{table}
In positive-unlabelled learning the standard procedure~\cite{elkan2008pu} trains a classifier on the observed labels and corrects it afterwards. The novelty-detection classifier is such a classifier: it treats a coarse label as a class of its own, and so estimates $P(\text{label}=n\mid x)$. The Elkan--Noto correction uses a pooled count estimate of $\hat q$ (over superclasses with two or more classes; within $0.011$ of Arbor's, which also counts single-class ones). In every superclass that has a class without fine-grained labels, it gives $p^*(\ell\mid x)=P(\text{label }\ell\mid x)/(1-\hat q)$ for each known class and the novelty mass $\max\{0,\,P(\text{label }y\mid x)-\hat q\sum_\ell p^*(\ell\mid x)\}$ on the superclass. Within each such superclass the corrected known-class mass is capped at the superclass's total probability and the rest goes to the superclass, so the posterior needs no renormalisation; the depth prior reads it out at the same operating point. Arbor fits the same likelihood during training (forward correction). \Cref{tab:pu} gives the results. The correction leaves the known-class argmax unchanged within a superclass. Read from the novelty-detection classifier's leaves (\cref{tab:merged}), known-class accuracy is $6$--$11$ points below Arbor's; read from the corrected posterior, whose per-superclass cap can change the argmax across superclasses (\cref{fig:gains}a), it is $3.7$--$11.0$ points below. The Elkan--Noto correction detects novel classes well, but its known-class predictions come from a classifier trained on the fine-grained labels alone. Over the $18$ settings of \cref{fig:gains}a, Arbor's known-class accuracy exceeds the correction's by $1.9$--$15.3$ points, and its novelty score separates novel from known classes better than the correction's in $15$ (AUROC; within NABirds images on the merge).

\paragraph{A two-stage baseline.} The two-stage baseline keeps that detector and takes the known classes from the classification baseline with the highest validation accuracy (the per-level classifiers on Cars and CIFAR-100, the conditional softmax on CUB, the flat softmax on Aircraft). It predicts that classifier's leaf, or the leaf's superclass when the Elkan--Noto correction's novelty mass there exceeds a threshold, set to the same operating point and swept in \cref{fig:curves}. It keeps that classifier's known-class accuracy but recognises fewer novel classes than Arbor on all four benchmarks, with more incorrect predictions over all test images (\cref{tab:pu}); it is less exact than Arbor on Cars, CUB and CIFAR-100 and $0.5$ points more exact on Aircraft.
\paragraph{Details of \cref{fig:gains}.} Panel (a) covers $18$ settings: the four small benchmarks under each of four protocols (the main protocol, $50\%$ coarse labels, two novel classes per superclass and class-dependent coarsening), plus iNat-2019 and the CUB--NABirds merge. The correction's known-class accuracy is read from the corrected posterior. On iNat-2019, where the correction was not run, it is the known-class accuracy of hierarchical novelty detection (\cref{tab:merged}). Panel (b) uses a fixed configuration of the linear classifier (learning rate $\times10$, $\rho=0.01$), iNat-2019 at $90$, $95$ and $98\%$ coarse labels and the other benchmarks at $50$ and $90\%$; three seeds, except a single seed for the linear classifier on iNat-2019 at $90\%$.

\section{Class-dependent coarsening}\label{app:varq}
Each known class is coarsened with its own probability $q_\ell$, drawn uniformly from $[0.80,0.98]$ (mean $0.89$), with the novel classes of the main protocol; Arbor keeps one pooled $\hat q$. \Cref{tab:varq} gives the main methods (hierarchical CE, soft labels and HypStructure were also run, with no change to the conclusions). \Cref{thm:pooled}(b) predicts that a class coarsened more often than $\hat q$ loses probability to the virtual novel class, by $(q_\ell-\hat q)/(1-\hat q)$. \Cref{tab:theorytest} tests this within superclasses with a virtual novel class. On every benchmark, the share of a known class's test images predicted as the superclass rises from the third of classes with the lowest $q_\ell$ to the third with the highest, while the same classifier without the virtual novel class predicts the superclass for at most $9\%$ of them in every third. The measured probability of the virtual novel class rises by a third to three fifths of the predicted amount, and has a floor of $0.11$--$0.40$ where the theorem predicts $0$. The floor arises because the virtual novel class also receives probability from uncertain known-class images, which the population optimum does not model. Novelty detection rises the same way, as \cref{thm:pooled}(a) implies for any model whose superclass probability is fitted to coarse labels.
\begin{table}[t]\centering\scriptsize\setlength{\tabcolsep}{3pt}
\resizebox{\columnwidth}{!}{\begin{tabular}{llcccc}
\toprule
third of $q_\ell$: low / middle / high & & Cars & CUB & CIFAR-100 & Aircraft \\
\midrule
virtual-novel-class probability & predicted & 0 / .09 / .60 & 0 / .09 / .58 & 0 / .08 / .58 & 0 / .07 / .59 \\
 & measured & .40 / .50 / .72 & .24 / .36 / .51 & .11 / .22 / .46 & .38 / .50 / .59 \\
superclass predictions & Arbor & 36 / 50 / 80 & 20 / 43 / 68 & 7 / 28 / 79 & 43 / 59 / 76 \\
 & novelty detection & 21 / 30 / 53 & 21 / 36 / 56 & 18 / 33 / 56 & 26 / 37 / 48 \\
 & no virtual novel class & 0 / 0 / 0 & 3 / 6 / 9 & 0 / 0 / 0 & 3 / 1 / 3 \\
\bottomrule
\end{tabular}}
\caption{\textbf{Testing \cref{thm:pooled}(b)} with coarsening probabilities that differ by class ($q_\ell\sim U[0.80,0.98]$), known classes split into thirds by $q_\ell$; superclasses with a virtual novel class, three seeds. Predicted: $\max\{0,(q_\ell-\hat q)/(1-\hat q)\}$. Superclass predictions: share of the class's test images (\%), for Arbor, for novelty detection, and for Arbor's classifier without the virtual novel class.}\label{tab:theorytest}\end{table}
In known-class accuracy, classes with $q_\ell\ge0.95$ lose $6$--$13$ points against the same classifier without the virtual novel class, and classes with $q_\ell<0.85$ lose $0$--$5$. This is a training effect; the population optimum, which keeps the leaf ratios of the partial-label likelihood, does not predict it. The single softmax with per-class $\hat q_\ell$ in every superclass (\cref{tab:varq}) recovers the classes with high $q_\ell$ but loses those with low $q_\ell$, and recognises far fewer novel classes. Arbor's known-class accuracy is within $1.5$ points of the most accurate classification baseline on Cars and Aircraft, and $4.6$ and $5.3$ points lower on CUB and CIFAR-100.
\begin{table*}[!t]\centering\scriptsize\setlength{\tabcolsep}{4pt}\resizebox{0.8\textwidth}{!}{\begin{tabular}{lcccccccc}
\toprule
 & \multicolumn{2}{c}{Cars} & \multicolumn{2}{c}{CUB} & \multicolumn{2}{c}{CIFAR-100} & \multicolumn{2}{c}{Aircraft} \\
\cmidrule(lr){2-3}\cmidrule(lr){4-5}\cmidrule(lr){6-7}\cmidrule(lr){8-9}
method & K & N & K & N & K & N & K & N \\
\midrule
Flat + HSC & .824 & 6 & .730 & 17 & .865 & 31 & .763 & 5 \\
Conditional & .788 & 17 & .748 & 18 & .860 & 55 & .775 & 9 \\
Per-node sigmoids & .761 & 12 & .732 & 20 & .872 & 60 & .727 & 22 \\
Per-level & .822 & 5 & .718 & 13 & .873 & 38 & .764 & 5 \\
\midrule
Uniform target & .778 & 48 & .645 & 35 & .686 & 76 & .778 & 37 \\
Novelty det. & .714 & 71 & .638 & 61 & .780 & 82 & .681 & 55 \\
\midrule
Arbor, no virtual novel class & .845 & 0 & .749 & 1 & .861 & 1 & .784 & 1 \\
single softmax, per-class $\hat q_\ell$ everywhere & .809 & 60 & .668 & 40 & .850 & 30 & .748 & 44 \\
\textbf{Arbor} & .809 & 86 & .702 & 76 & .819 & 76 & .761 & 78 \\
\midrule
\multicolumn{9}{l}{\textit{per-class known accuracy (\%) by the class's own $q_\ell$, superclasses with a novel class}} \\
Arbor, no virtual novel class, $q_\ell<0.85$ & \multicolumn{2}{c}{91} & \multicolumn{2}{c}{88} & \multicolumn{2}{c}{92} & \multicolumn{2}{c}{71} \\
Arbor, no virtual novel class, $0.85\le q_\ell<0.95$ & \multicolumn{2}{c}{79} & \multicolumn{2}{c}{68} & \multicolumn{2}{c}{90} & \multicolumn{2}{c}{60} \\
Arbor, no virtual novel class, $q_\ell\ge 0.95$ & \multicolumn{2}{c}{62} & \multicolumn{2}{c}{43} & \multicolumn{2}{c}{67} & \multicolumn{2}{c}{35} \\
single softmax, per-class $\hat q_\ell$ everywhere, $q_\ell<0.85$ & \multicolumn{2}{c}{77} & \multicolumn{2}{c}{68} & \multicolumn{2}{c}{87} & \multicolumn{2}{c}{46} \\
single softmax, per-class $\hat q_\ell$ everywhere, $0.85\le q_\ell<0.95$ & \multicolumn{2}{c}{76} & \multicolumn{2}{c}{62} & \multicolumn{2}{c}{87} & \multicolumn{2}{c}{61} \\
single softmax, per-class $\hat q_\ell$ everywhere, $q_\ell\ge 0.95$ & \multicolumn{2}{c}{69} & \multicolumn{2}{c}{51} & \multicolumn{2}{c}{84} & \multicolumn{2}{c}{50} \\
\textbf{Arbor}, $q_\ell<0.85$ & \multicolumn{2}{c}{86} & \multicolumn{2}{c}{85} & \multicolumn{2}{c}{92} & \multicolumn{2}{c}{68} \\
\textbf{Arbor}, $0.85\le q_\ell<0.95$ & \multicolumn{2}{c}{71} & \multicolumn{2}{c}{64} & \multicolumn{2}{c}{87} & \multicolumn{2}{c}{55} \\
\textbf{Arbor}, $q_\ell\ge 0.95$ & \multicolumn{2}{c}{56} & \multicolumn{2}{c}{31} & \multicolumn{2}{c}{54} & \multicolumn{2}{c}{22} \\
\bottomrule
\end{tabular}
}
\caption{\textbf{Coarsening probabilities that differ by class}, $q_\ell\sim U[0.80,0.98]$, one novel class per superclass. K: known-class accuracy; N: novel-class accuracy (\%). Bottom: per-class known accuracy by the class's own $q_\ell$, over classes and seeds. Three seeds.}\label{tab:varq}\end{table*}

\section{Count-informed inference for the baselines}\label{app:countaware}
This inference rule gives a classification baseline the information that Arbor's loss reads from the training labels, namely which superclasses contain a class without fine-labelled training images. Leaf predictions in such a superclass fall back to it unless the leaf confidence exceeds a threshold $\theta_u$. Predictions elsewhere use the classifier's usual fallback threshold $\theta$, set so that fine-labelled validation images keep a mean specificity of $0.9$. Among the admissible pairs $(\theta,\theta_u)$, the one with the highest superclass rate on coarse-labelled validation images is chosen. Arbor is then set on validation to the baseline's exact-class rate, with its own depth prior or with the same count-informed rule (Arbor$^\dagger$). The rule raises the classification baselines' novel-class accuracy by $6$--$51$ points on Cars, CIFAR-100 and Aircraft and by at most $12$ on CUB. On the seed means, only the conditional softmax with this rule is ahead of Arbor's depth prior, on CIFAR-100 ($87$ vs $79\%$). With the same rule, Arbor recognises more novel classes than the baseline in $98$ of the $108$ combinations of benchmark, seed and baseline (\cref{tab:countaware_main}); the exceptions are the conditional softmax on CIFAR-100 in all three seeds ($-2.5$ to $-4.5$ points) and seven baselines on one Aircraft seed ($-0.6$ to $-4.1$); for these, the thresholds chosen on validation predict the superclass for every image in a superclass with a novel class.
\begin{table*}[!t]\centering\scriptsize\setlength{\tabcolsep}{4pt}\resizebox{0.82\textwidth}{!}{\begin{tabular}{lcccccccccccc}
\toprule
 & \multicolumn{3}{c}{Cars} & \multicolumn{3}{c}{CUB} & \multicolumn{3}{c}{CIFAR-100} & \multicolumn{3}{c}{Aircraft} \\
\cmidrule(lr){2-4}\cmidrule(lr){5-7}\cmidrule(lr){8-10}\cmidrule(lr){11-13}
count-informed baseline & base. & Arbor & Arbor$^\dagger$ & base. & Arbor & Arbor$^\dagger$ & base. & Arbor & Arbor$^\dagger$ & base. & Arbor & Arbor$^\dagger$ \\
\midrule
Flat + HSC & 66 / 35 & 64 / \textbf{81} & 64 / \textbf{81} & 58 / 21 & 61 / \textbf{70} & 61 / \textbf{70} & 68 / 30 & 70 / 79 & 69 / \textbf{82} & 66 / 53 & 65 / 71 & 65 / \textbf{71} \\
Conditional & 64 / 40 & 64 / 80 & 62 / \textbf{80} & 60 / 20 & 61 / \textbf{69} & 61 / \textbf{69} & 68 / \textbf{87} & 69 / 79 & 68 / 83 & 64 / 47 & 62 / 77 & 62 / \textbf{78} \\
Per-node sigmoids & 63 / 35 & 65 / \textbf{79} & 64 / \textbf{79} & 59 / 17 & 61 / \textbf{68} & 61 / \textbf{68} & 67 / 64 & 68 / 80 & 68 / \textbf{83} & 64 / 48 & 63 / 77 & 63 / \textbf{77} \\
Per-level & 62 / 43 & 64 / \textbf{81} & 63 / \textbf{81} & 61 / 15 & 63 / \textbf{66} & 63 / \textbf{66} & 68 / 63 & 70 / 79 & 70 / \textbf{82} & 65 / 57 & 63 / 77 & 63 / \textbf{77} \\
Hier.\ CE & 62 / 40 & 64 / \textbf{80} & 63 / 80 & 57 / 19 & 60 / \textbf{72} & 60 / \textbf{72} & 69 / 53 & 69 / 80 & 69 / \textbf{83} & 66 / 46 & 62 / 78 & 62 / \textbf{79} \\
Soft labels & 64 / 39 & 66 / 78 & 64 / \textbf{79} & 58 / 24 & 63 / \textbf{66} & 63 / \textbf{66} & 68 / 39 & 70 / 79 & 69 / \textbf{82} & 64 / 51 & 63 / 77 & 63 / \textbf{77} \\
HypStructure & 63 / 35 & 63 / \textbf{82} & 62 / \textbf{82} & 56 / 15 & 58 / \textbf{76} & 58 / \textbf{76} & 68 / 33 & 69 / 80 & 69 / \textbf{83} & 64 / 52 & 63 / 76 & 63 / \textbf{76} \\
\midrule
Uniform target & 61 / 49 & 63 / 82 & 62 / \textbf{83} & 54 / 25 & 56 / \textbf{78} & 56 / \textbf{78} & 61 / 68 & 63 / 83 & 62 / \textbf{87} & 65 / 54 & 64 / 72 & 64 / \textbf{73} \\
Novelty det. & 55 / 68 & 53 / 88 & 53 / \textbf{88} & 53 / 45 & 52 / \textbf{84} & 52 / \textbf{84} & 60 / 78 & 63 / 83 & 61 / \textbf{86} & 49 / 57 & 47 / 81 & 57 / \textbf{89} \\
\bottomrule
\end{tabular}
}
\caption{\textbf{Baselines given the count information Arbor's loss uses, in the main protocol} ($90\%$ coarse labels, one novel class per superclass; \%). Each cell: exact-class rate on known-class images / novel-class accuracy. Baseline: count-informed inference rule with both thresholds chosen on validation. Arbor: its depth-prior threshold; Arbor$^\dagger$: the same count-informed rule; both set on validation to that baseline's exact-class rate. Bold: highest novel-class accuracy of the three, by unrounded value. Three seeds.}\label{tab:countaware_main}\end{table*}

\section{Additional protocols: 50\% coarse labels, a random novel class, two novel classes}\label{app:moreprotocols}
Across these protocols the novelty baselines exceed Arbor in novel-class accuracy only on CIFAR-100, at $2$--$5$ points lower known-class accuracy.
\Cref{tab:q05} repeats \cref{tab:merged} with $50\%$ coarse labels, \cref{tab:randnovel} draws the novel class of each superclass at random per seed, and \cref{tab:twonever} places two novel classes in every superclass of three or more. With $50\%$ coarse labels, Arbor's known-class accuracy is $1.0$--$2.5$ points below the most accurate classification baseline's and it recognises $83$--$90\%$ of novel-class images, against at most $41\%$ for the classification baselines on Cars, CUB and Aircraft and $86\%$ for the per-node sigmoids on CIFAR-100, where hierarchical novelty detection recognises $91\%$. With a random novel class, Arbor recognises more novel classes than every classification baseline run on all four benchmarks, by $18$--$71$ points, including CIFAR-100, where the conditional softmax falls from $81$ to $51\%$; hierarchical novelty detection recognises more than Arbor only on CIFAR-100 ($84\%$), at $5$ points lower known-class accuracy. With two novel classes, Arbor's known-class accuracy is $0.9$ points above the most accurate classification baseline's on Cars and $0.7$--$3.7$ points below it elsewhere, and it recognises $87$--$92\%$ of novel-class images, more than every method on every benchmark.
\begin{table*}[t]\centering\scriptsize\setlength{\tabcolsep}{3pt}
\resizebox{0.82\textwidth}{!}{\begin{tabular}{lcccccccccccc}
\toprule
 & \multicolumn{3}{c}{Cars} & \multicolumn{3}{c}{CUB} & \multicolumn{3}{c}{CIFAR-100} & \multicolumn{3}{c}{Aircraft} \\
\cmidrule(lr){2-4}\cmidrule(lr){5-7}\cmidrule(lr){8-10}\cmidrule(lr){11-13}
method & novel$\uparrow$ & leaf$\downarrow$ & known$\uparrow$ & novel$\uparrow$ & leaf$\downarrow$ & known$\uparrow$ & novel$\uparrow$ & leaf$\downarrow$ & known$\uparrow$ & novel$\uparrow$ & leaf$\downarrow$ & known$\uparrow$ \\
\midrule
Flat softmax + HSC~\cite{goren2024hsc} & 4 & 86 & \textbf{.917} & 13 & 82 & .903 & 11 & 79 & \textbf{.919} & 8 & 83 & .840 \\
Conditional softmax~\cite{redmon2017yolo9000,valmadre2022hierarchical} & 23 & 66 & .896 & 41 & 54 & .899 & 83 & 8 & .915 & 11 & 80 & .829 \\
Per-node sigmoids~\cite{valmadre2022hierarchical} & 12 & 76 & .891 & 29 & 68 & .887 & 86 & 8 & \textbf{.919} & 27 & 64 & .831 \\
Per-level + fallback~\cite{park2025hcast,garg2022haf,chu2026hieruq} & 6 & 80 & .915 & 18 & 76 & .908 & 34 & 55 & \textbf{.919} & 8 & 80 & .839 \\
Hierarchical CE~\cite{bertinetto2020making} & 11 & 80 & .913 & 26 & 70 & .906 & 22 & 69 & .917 & 10 & 81 & \textbf{.843} \\
Soft labels~\cite{bertinetto2020making} & 7 & 85 & .912 & 13 & 82 & .903 & 8 & 82 & .918 & 8 & 83 & .840 \\
HypStructure~\cite{zeng2024hypstructure} & 4 & 86 & .908 & 21 & 74 & \textbf{.909} & 16 & 73 & .915 & 6 & 83 & .829 \\
\midrule
Uniform target & 73 & 19 & .892 & 79 & 18 & .869 & 88 & \textbf{5} & .901 & 39 & 51 & .827 \\
Hierarchical novelty detection~\cite{lee2018hierarchical} & 87 & 8 & .836 & 85 & 13 & .833 & \textbf{91} & 6 & .890 & 74 & 20 & .779 \\
\midrule
\rowcolor{ourspastel} \textbf{Arbor (ours)} & \textbf{87} & \textbf{7} & .892 & \textbf{90} & \textbf{8} & .886 & 89 & \textbf{5} & .909 & \textbf{83} & \textbf{10} & .833 \\
\bottomrule
\end{tabular}
}
\caption{\textbf{$50\%$ coarse labels}, one novel class per superclass; layout as in \cref{tab:merged} (novel and leaf: novel-class images assigned to their superclass or to a leaf, \%; known: known-class accuracy). Three seeds. Bold: column best (accuracies: ties at the precision shown; percentages: within half a point).}\label{tab:q05}\end{table*}
\begin{table*}[t]\centering\scriptsize\setlength{\tabcolsep}{3pt}
\resizebox{0.82\textwidth}{!}{\begin{tabular}{lcccccccccccc}
\toprule
 & \multicolumn{3}{c}{Cars} & \multicolumn{3}{c}{CUB} & \multicolumn{3}{c}{CIFAR-100} & \multicolumn{3}{c}{Aircraft} \\
\cmidrule(lr){2-4}\cmidrule(lr){5-7}\cmidrule(lr){8-10}\cmidrule(lr){11-13}
method & novel$\uparrow$ & leaf$\downarrow$ & known$\uparrow$ & novel$\uparrow$ & leaf$\downarrow$ & known$\uparrow$ & novel$\uparrow$ & leaf$\downarrow$ & known$\uparrow$ & novel$\uparrow$ & leaf$\downarrow$ & known$\uparrow$ \\
\midrule
Flat softmax + HSC~\cite{goren2024hsc} & 5 & 87 & .818 & 11 & 86 & .714 & 27 & 65 & .884 & 9 & 83 & \textbf{.794} \\
Conditional softmax~\cite{redmon2017yolo9000,valmadre2022hierarchical} & 11 & 84 & .806 & 18 & 80 & .744 & 51 & 42 & .874 & 15 & 74 & .790 \\
Per-node sigmoids~\cite{valmadre2022hierarchical} & 10 & 77 & .797 & 11 & 87 & .740 & 49 & 44 & .888 & 18 & 74 & .744 \\
Per-level + fallback~\cite{park2025hcast,garg2022haf,chu2026hieruq} & 6 & 83 & \textbf{.827} & 11 & 84 & \textbf{.745} & 42 & 49 & \textbf{.905} & 9 & 81 & .786 \\
\midrule
Uniform target & 35 & 61 & .764 & 21 & 79 & .647 & 66 & 30 & .749 & 46 & 48 & .787 \\
Hierarchical novelty detection~\cite{lee2018hierarchical} & 60 & 37 & .731 & 60 & 37 & .617 & \textbf{84} & \textbf{14} & .813 & 48 & 50 & .712 \\
\midrule
\rowcolor{ourspastel} \textbf{Arbor (ours)} & \textbf{81} & \textbf{16} & .821 & \textbf{80} & \textbf{18} & .734 & 69 & 25 & .864 & \textbf{75} & \textbf{19} & .789 \\
\bottomrule
\end{tabular}
}
\caption{\textbf{The novel class drawn at random per seed}, $90\%$ coarse labels; layout as in \cref{tab:merged} (novel and leaf: novel-class images assigned to their superclass or to a leaf, \%; known: known-class accuracy). Three seeds, each with its own set of novel classes. Hierarchical cross-entropy, soft labels and HypStructure were not run in this protocol. Bold: column best (accuracies: ties at the precision shown; percentages: within half a point).}\label{tab:randnovel}\end{table*}
\begin{table*}[t]\centering\scriptsize\setlength{\tabcolsep}{3pt}
\resizebox{0.82\textwidth}{!}{\begin{tabular}{lcccccccccccc}
\toprule
 & \multicolumn{3}{c}{Cars} & \multicolumn{3}{c}{CUB} & \multicolumn{3}{c}{CIFAR-100} & \multicolumn{3}{c}{Aircraft} \\
\cmidrule(lr){2-4}\cmidrule(lr){5-7}\cmidrule(lr){8-10}\cmidrule(lr){11-13}
method & novel$\uparrow$ & leaf$\downarrow$ & known$\uparrow$ & novel$\uparrow$ & leaf$\downarrow$ & known$\uparrow$ & novel$\uparrow$ & leaf$\downarrow$ & known$\uparrow$ & novel$\uparrow$ & leaf$\downarrow$ & known$\uparrow$ \\
\midrule
Flat softmax + HSC~\cite{goren2024hsc} & 6 & 84 & .833 & 21 & 76 & .752 & 19 & 73 & .893 & 11 & 79 & .763 \\
Conditional softmax~\cite{redmon2017yolo9000,valmadre2022hierarchical} & 14 & 79 & .823 & 27 & 70 & .769 & 75 & 19 & .893 & 21 & 67 & \textbf{.784} \\
Per-node sigmoids~\cite{valmadre2022hierarchical} & 13 & 72 & .815 & 27 & 72 & .758 & 44 & 48 & .906 & 30 & 59 & .743 \\
Per-level + fallback~\cite{park2025hcast,garg2022haf,chu2026hieruq} & 6 & 83 & .840 & 19 & 75 & \textbf{.778} & 26 & 65 & \textbf{.919} & 11 & 77 & .762 \\
Hierarchical CE~\cite{bertinetto2020making} & 6 & 84 & .792 & 20 & 78 & .742 & 32 & 60 & .890 & 23 & 64 & .764 \\
Soft labels~\cite{bertinetto2020making} & 7 & 81 & .830 & 21 & 76 & .752 & 25 & 66 & .895 & 12 & 75 & .771 \\
HypStructure~\cite{zeng2024hypstructure} & 5 & 86 & .817 & 21 & 75 & .743 & 14 & 79 & .894 & 14 & 76 & .744 \\
\midrule
Uniform target & 84 & 12 & .796 & 63 & 36 & .690 & 68 & 28 & .750 & 83 & 10 & .762 \\
Hierarchical novelty detection~\cite{lee2018hierarchical} & 68 & 31 & .740 & 66 & 33 & .663 & 81 & 16 & .843 & 52 & 45 & .695 \\
\midrule
\rowcolor{ourspastel} \textbf{Arbor (ours)} & \textbf{92} & \textbf{5} & \textbf{.849} & \textbf{88} & \textbf{11} & .741 & \textbf{87} & \textbf{10} & .888 & \textbf{89} & \textbf{4} & .778 \\
\bottomrule
\end{tabular}
}
\caption{\textbf{Two novel classes per superclass} of three or more classes and none in superclasses of two, $90\%$ coarse labels; only the superclass is identified (\cref{sec:theory}). Layout as in \cref{tab:merged} (novel and leaf: novel-class images assigned to their superclass or to a leaf, \%; known: known-class accuracy). Three seeds. Bold: column best (accuracies: ties at the precision shown; percentages: within half a point).}\label{tab:twonever}\end{table*}

\section{A large taxonomy: iNaturalist-2019}\label{app:inat}
On iNat-2019 the mixed-granularity protocol has one novel class per genus and about $19$ fine-grained labels per known class. \Cref{tab:inat} decomposes Arbor's known-class gap with diagnostic variants on one seed, compared with Arbor on the same seed. Without the virtual novel class, the same classifier gains $1.0$ point of known-class accuracy and recognises $20\%$ of novel-class images. A larger embedding ($k=256$), twice the training epochs, free prototypes and a linear classifier over the nodes with Arbor's loss and gate do not close the gap to the per-level classifiers ($5.0$ points in this seed); the best of them, the linear classifier, is $1.3$ points more accurate at $66\%$ novel-class accuracy. The gap is therefore attributable to the node classifier rather than to the virtual novel class. \Cref{fig:gains}b plots Arbor's known-class accuracy minus the linear classifier's against the number of fine-grained labels per class, which is four to seven on Cars, CUB and Aircraft and about $42$ on CIFAR-100. Over all test images, at the same operating point, the classification baselines make $22.7$--$27.5\%$ incorrect predictions (\cref{tab:inat} shows two of them), Arbor $28.1\%$ and the novelty baselines $33.8$ and $39.4\%$; novel-class images are about one test image in thirteen.
\begin{table}[t]\centering\scriptsize\setlength{\tabcolsep}{3.5pt}
\begin{tabular}{lcccc}
\toprule
method & known & novel & leaf & incorrect, all \\
\midrule
Per-level + fallback & .646 & 30 & 65 & 23.7 \\
Flat softmax + HSC & .639 & 34 & 61 & 22.7 \\
Uniform target & .467 & 46 & 52 & 33.8 \\
Hierarchical novelty detection & .441 & 43 & 55 & 39.4 \\
\rowcolor{ourspastel} \textbf{Arbor} & .594 & 68 & 31 & 28.1 \\
\midrule
\multicolumn{5}{l}{\textit{diagnostic variants, seed 0}} \\
Arbor, no virtual novel class & .605 & 20 & 72 & -- \\
Arbor, embedding $k=256$ & .592 & 69 & 30 & -- \\
Arbor, $80$ epochs & .566 & 54 & 42 & -- \\
Arbor, free prototypes & .604 & 70 & 29 & -- \\
Arbor's loss and gate, linear classifier & .608 & 66 & 33 & 27.0 \\
Arbor, same seed & .595 & 70 & 29 & 27.9 \\
\bottomrule
\end{tabular}
\caption{\textbf{iNaturalist-2019, mixed-granularity labels} ($90\%$ coarse, one novel class per genus). Known: known-class accuracy; novel and leaf: novel-class images assigned to their genus or to a leaf (\%); incorrect, all: predictions over all test images that are neither the true class nor an ancestor (\%), at the same operating point. Top: three seeds.}\label{tab:inat}\end{table}

\section{A merged dataset: CUB and NABirds}\label{app:merge}
We merge two datasets whose granularities differ. CUB-200-2011 gives fine-grained labels; NABirds~\cite{vanhorn2015nabirds} covers the same North American birds and more, and we give its images only their family-level label. We use NABirds' own taxonomy, with its order-level groups at the top and its family-level groups as superclasses, and keep the $43$ superclasses that contain a CUB species. The $140$ CUB species found in NABirds by name are the known classes; the $108$ NABirds species without a CUB counterpart are the novel classes, in $19$ of the superclasses. We remove $243$ NABirds images that near-duplicate a CUB image (cosine similarity of DINOv2 features above $0.95$), leaving $8{,}355$ CUB and $31{,}304$ NABirds images. Splits are $70/10/20$ by species over both sources; every NABirds training and validation image carries only its superclass label, and test images keep their species. A known class is coarse with probability equal to its share of NABirds images, $0.34$--$0.88$ (pooled $0.69$), so the coarsening is class-dependent and set by the data, and each known class keeps about $42$ fine-grained labels. The count estimate gives $\hat q=0.64$, low because the novel classes come from the coarse source only.

\paragraph{Source.} Every novel test image is from NABirds, so a method could detect novel classes by detecting NABirds photographs. A logistic probe on the frozen features separates the two sources only weakly (balanced accuracy $0.65$, and $0.63$ within a species), and we score novelty within NABirds test images alone.

\paragraph{Results.} \Cref{tab:merge} gives the results. Its AUROC column scores, for Arbor, novelty detection and the Elkan--Noto correction, the probability of the virtual novel class of the image's true superclass; for the two-stage baseline, the corrected novelty mass of the predicted leaf's superclass; for the other methods, one minus the largest leaf probability. The Elkan--Noto correction keeps novelty detection's known-class predictions, and the two-stage baseline the per-level classifiers' (\cref{app:pu}). Within NABirds images, Arbor's AUROC is within $.01$ of the Elkan--Noto correction's and the two-stage baseline's, and above every other method's. At each of six baselines' exact-class rates (the per-level, conditional, flat and soft-label classifiers, the uniform target and hierarchical novelty detection), Arbor assigns $79$--$95\%$ of novel-class images to their superclass, against $33$--$93\%$. It makes fewer incorrect predictions over all test images than each of them except the conditional softmax ($8.5$ vs $5.2\%$) and the uniform target ($5.6$ vs $4.2\%$). Against the two-stage baseline, each at its own operating point, neither method dominates. Arbor recognises $94.1\%$ of novel species against $92.6$; the two-stage baseline makes $4.9\%$ incorrect predictions over all test images against Arbor's $5.4$ and is $4.4$ points more accurate on known classes ($.883$ vs $.839$; $3.3$ points on CUB test images and $4.9$ on NABirds test images). With a linear classifier or free prototypes in place of the tree, Arbor's loss and gate reach $.825$ and $.835$ known-class accuracy and AUROC $.832$ and $.818$. As \cref{thm:pooled}(b) predicts, within superclasses with a novel class the share of known-class images predicted as the superclass rises with $q_\ell$, from $36\%$ in the third of classes with the lowest $q_\ell$ to $68\%$ in the highest; the held-out likelihood selects $q=0.5$, below the pooled $0.69$, consistent with \cref{thm:pooled}(c).
\begin{table}[t]\centering\scriptsize\setlength{\tabcolsep}{3pt}
\resizebox{\columnwidth}{!}{\begin{tabular}{lccccc}
\toprule
method & known & known: CUB / NABirds & novel & exact & AUROC \\
\midrule
\rowcolor{ourspastel} \textbf{Arbor} & .839 & 87.8 / 82.1 & \textbf{94.1} & 49.6 & .818 \\
Elkan--Noto correction & .799 & 84.8 / 77.7 & 93.2 & 48.3 & \textbf{.822} \\
two-stage (best classifier + Elkan--Noto) & \textbf{.883} & \textbf{91.1} / \textbf{87.0} & 92.6 & 50.0 & \textbf{.822} \\
hierarchical novelty detection & .799 & 84.8 / 77.7 & 92.2 & 46.2 & .775 \\
uniform target & .755 & 84.7 / 71.3 & 93.2 & 50.3 & .767 \\
\midrule
per-level + fallback & \textbf{.883} & \textbf{91.1} / \textbf{87.0} & 40.2 & \textbf{69.1} & .627 \\
conditional softmax & .873 & 90.4 / 86.0 & 83.5 & 60.7 & .759 \\
flat softmax + HSC & .872 & 90.8 / 85.6 & 33.1 & 67.7 & .491 \\
soft labels & .870 & 90.4 / 85.4 & 49.5 & 66.6 & .636 \\
per-node sigmoids & .866 & 89.3 / 85.4 & 40.9 & 66.1 & .596 \\
hierarchical CE & .863 & 90.2 / 84.5 & 55.6 & 66.2 & .663 \\
HypStructure & .873 & 90.8 / 85.8 & 36.7 & 66.8 & .511 \\
\bottomrule
\end{tabular}}
\caption{\textbf{The CUB--NABirds merge} (CUB fine-grained, NABirds coarse; $140$ known and $108$ novel species; three seeds). Known: known-class accuracy (fraction), overall and by test-image source (CUB / NABirds, \%). Novel, exact: novel-class accuracy and exact-class rate (\%) at the same operating point. AUROC: novelty score, novel- vs known-class images within NABirds test images (scores defined in \cref{app:merge}). Bold: column best.}\label{tab:merge}\end{table}
\section{Naturally coarse labels: iNaturalist}\label{app:natural}
We merge two sources whose coarsening we did not design. The fine source is research-grade iNaturalist observations, from 2019 onwards, of iNat-2019's species. The coarse source is iNaturalist observations that the community could identify only to genus. All images are research-grade or genus-level observations from 2019 onwards with openly licensed photos, so known and novel test images come from the same source and period. We keep the $55$ of iNat-2019's $72$ genera whose species names are all still current, so that a renamed known species can never count as novel. The taxonomy lists the $778$ known species and every other species of these genera with at least $100$ research-grade observations ($2{,}904$ novel species, $3{,}682$ leaves). Each known species has up to $130$ research-grade observations (median $130$), split $100/10/20$ images per species into training, validation and test. Each genus has up to $1{,}000$ genus-only observations ($90/10$ training and validation), whose species is unknown. Each novel species has $15$ research-grade test observations. The coarsening here can depend on the image: observations that the community cannot resolve may be harder photographs. Hierarchical cross-entropy ran out of memory at $3{,}682$ classes and is not reported.

Arbor's count-based $\hat q$ assumes that every listed species has the same number of images, and here clamps to $0$, so every genus-only label is explained by the gate. We also ran the Elkan--Noto label-frequency estimate~\cite{elkan2008pu}: a logistic regression separates fine- from coarse-labelled training images, and $\hat q=1-\bar e$ with $\bar e$ its mean score on fine-labelled validation images ($\hat q=0.32$--$0.33$ across seeds). \Cref{tab:natural} gives the results. The classification baselines recognise $18$--$50\%$ of novel species and hierarchical novelty detection $56\%$ at $.686$ known-class accuracy; with either estimate, Arbor recognises $48$--$49\%$ at $.712$--$.722$ known-class accuracy, close to the uniform target ($.727$, $49\%$) and below soft labels in both. Selecting $q$ on validation rather than estimating it from counts gives no improvement. Fine-grained validation accuracy, which does not depend on novel-class images, selects $q=0.9$, where Arbor recognises $3\%$ of novel species: the overestimate of \cref{thm:pooled}(a), which clips the virtual novel class to zero wherever $c(x)<\hat q$. The held-out likelihood ranks $q=0.5$ highest, consistent with \cref{thm:pooled}(c); there Arbor recognises $41\%$. Since the coarsening here can depend on the image, \cref{eq:coarsen} does not hold, and we expect the gate's score, which separates fine- from coarse-labelled images, to reflect in part the difficulty of the photograph rather than only the novelty of the species.
\begin{table}[t]\centering\scriptsize\setlength{\tabcolsep}{4pt}
\begin{tabular}{lccc}
\toprule
method & known & novel & leaf \\
\midrule
Flat softmax + HSC & .760 & 31 & 52 \\
Conditional softmax & .752 & 43 & 43 \\
Per-node sigmoids & .704 & 18 & 81 \\
Per-level + fallback & .758 & 40 & 44 \\
Soft labels & .752 & 50 & 39 \\
HypStructure & .758 & 40 & 46 \\
\midrule
Uniform target & .727 & 49 & 38 \\
Hierarchical novelty detection & .686 & 56 & 33 \\
\midrule
\rowcolor{ourspastel} Arbor, count-based $\hat q$ ($=0$) & .712 & 49 & 40 \\
\rowcolor{ourspastel} Arbor, Elkan--Noto $\hat q$ & .722 & 48 & 40 \\
Arbor's loss and gate, linear classifier & .704 & 48 & 45 \\
\midrule
\multicolumn{4}{l}{\textit{Arbor, $q$ selected on validation}} \\
\quad $q=0.9$ (fine-labelled validation accuracy) & .732 & 3 & 71 \\
\quad $q=0.7$ & .727 & 22 & 59 \\
\quad $q=0.5$ (highest held-out likelihood) & .724 & 41 & 46 \\
\bottomrule
\end{tabular}
\caption{\textbf{Naturally coarse labels} (iNaturalist, $55$ genera). Known: accuracy on known-species test images; novel and leaf: novel species assigned to their genus or to a species (\%). Same operating point; three seeds. Bottom block: $q$ chosen by fine-labelled validation accuracy ($0.9$), the $q=0.7$ candidate, and $q=0.5$ with $\rho$ chosen on validation. Grid exceptions in \cref{app:data}.}\label{tab:natural}\end{table}

\section{Known classes with few fine-grained labels}\label{app:weak}
\begin{table}[t]\centering\scriptsize\setlength{\tabcolsep}{4pt}
\resizebox{\columnwidth}{!}{\begin{tabular}{lcccc}
\toprule
held class at $n_b{=}2$; siblings at $n_a{=}$ & 2 & 5 & 20 & all \\
\midrule
CIFAR-100 & 60 / 30 / 0 & 79 / 11 / 0 & 88 / 0 / 0 & 89 / 0 / 0 \\
Cars & 26 / 63 / 0 & 48 / 41 / 0 & 63 / 26 / 0 & 65 / 25 / 0 \\
CUB & 28 / 65 / 0 & 64 / 26 / 0 & 77 / 14 / 0 & 77 / 15 / 0 \\
Aircraft & 41 / 44 / 0 & 76 / 10 / 0 & 87 / 0 / 0 & 89 / 0 / 0 \\
Cars, CLIP & 24 / 68 / 0 & 61 / 31 / 0 & 71 / 20 / 0 & 74 / 18 / 0 \\
CUB, CLIP & 34 / 59 / 0 & 63 / 27 / 0 & 70 / 21 / 0 & 68 / 25 / 0 \\
\bottomrule
\end{tabular}
}
\caption{\textbf{Sibling misassignment against the siblings' number of fine-grained labels (\cref{prop:absorb}).} The held class keeps two fine-labelled images throughout; only its siblings' fine-grained labels are cut to $n_a$. Each cell: \% of the held class's test images assigned to a sibling / recognised / assigned to the superclass. Arbor ($\rho=0.03$), same operating point, three seeds; rows marked CLIP use OpenCLIP ViT-B/16 features.}
\label{tab:mechanism}\end{table}
Arbor's loss treats a class as known from its first fine-labelled image. A class with one or two fine-labelled images is then classified against siblings that keep all their fine-grained labels, and \cref{prop:absorb} predicts that its images are assigned to a sibling. \Cref{tab:mechanism} tests the lemma's prediction with the intervention it specifies: the held class keeps two fine-labelled images, and only its siblings' fine-grained labels are cut to $n_a$. The share of its test images assigned to a sibling rises with $n_a$ from $2$ to $20$ in all six rows and changes by at most $2.4$ points from $20$ to all. At balance ($n_a=n_b=2$) it is under half the rate with all sibling labels in five rows, and $0.68$ of it on CIFAR-100, and recognition of the two-image class rises from $0$--$25\%$ to $30$--$68\%$. A dependence on $n_b$ alone would give flat rows; every row rises with $n_a$. Arbor has no mechanism that keeps such a class on its superclass.

\section{Unseen classes}\label{app:unseen}
One leaf is held out of every superclass with three or more leaves, with no coarse labels, so the task is novelty detection, not hierarchical classification; the ideal prediction for a held-out class is its superclass. At the same operating point, Arbor assigns none of the unseen-class images to their superclass and $38$--$74\%$ to a leaf, against $30$--$61\%$ for the flat softmax; for a class absent from the taxonomy, any leaf is incorrect. The hyperbolic Busemann classifier, with internal prototypes near the origin, assigns them to their superclass most often ($73\%$ on CUB). Without coarse labels Arbor has no novelty gate, and every fine-grained label lowers the logit of the superclass node (\cref{app:theory}), so an unseen class is read as a sibling leaf.

\section{Released code of prior methods}\label{app:official}
The per-level row of the main tables is our reimplementation of the classifiers of H-CAST, HAF and HierUQ on the shared backbone. We also ran H-CAST's released code on CUB for two models: Hier-ViT, a DeiT-small~\cite{touvron2021deit} trained end to end from ImageNet-1k weights, and H-CAST proper, CAST-small~\cite{ke2024cast} with superpixel tokens and the global consistency term. Both use our $70$-superclass tree, our splits, best-epoch selection on our validation split, the species loss masked on coarse-labelled images and mixup off, and are scored with the same fallback rule and operating point as the reimplementation, over three seeds. With full labels, fine-grained accuracy is $.81$ (Hier-ViT) and $.86$ (H-CAST), against $.90$ for the DINOv2 baselines, which measures the backbone gap; at $90\%$ coarse labels it is $.35$ and $.52$, against $.73$ for the reimplementation and $.71$ for Arbor. With only the novel classes coarse, Hier-ViT assigns $52\%$ of images to the correct superclass and H-CAST $65\%$, with $37$ and $26\%$ leaf predictions; the reimplementation assigns $24\%$, and Arbor $86\%$ with $7\%$. The released code recognises more novel classes than our reimplementation of its classifier; both remain well below Arbor.

\paragraph{Free-grained recognition.} Free-grained hierarchical visual recognition~\cite{park2026freegrain} is the published method closest to our setting, and we ran its released code in its CUB configuration: H-ViT, a DeiT-small trained end to end from ImageNet-1k weights with its text-attribute contrastive loss on the captions it ships (CLIP~\cite{radford2021clip} ViT-B/32 text encoder), for $100$ epochs at batch $128$ with its learning-rate scaling. Its captions cover CUB's official training split only, so this comparison uses the official split: $5{,}394$ training images after a $10\%$ validation hold-out that selects the epoch, and the official test set of $5{,}794$. It uses our $70$-superclass tree, two coarse-label protocols and three seeds, with Arbor and the DINOv2 baselines on the same split and their embedding penalties chosen on validation (\cref{tab:offsplit_cmp}). At $90\%$ coarse labels it reaches $.488$ fine-grained accuracy, the level of a frozen flat softmax trained on the fine-labelled images alone ($.480$), against $.630$ for Arbor on this split. On novel classes it assigns $24.9\%$ of images to the correct superclass under our fallback ($62.6\%$ leaf predictions), $16.8\%$ under its own consistency rule ($79.0\%$) and $36.9\%$ under its confidence rule at threshold $0.9$ ($40.4\%$); Arbor reaches $84.6\%$ with $9.1\%$. This split leaves $2.7$ fine-labelled images per class at $90\%$ coarse labels, the regime of \cref{prop:absorb}: Arbor's known-class accuracy is within $0.7$ points of the per-level classifiers' and the conditional softmax's there, and it assigns $85\%$ of novel-class images to their superclass against $20\%$ for the per-level classifiers.

\begin{table}[t]\centering\scriptsize\setlength{\tabcolsep}{3pt}\resizebox{\columnwidth}{!}{\begin{tabular}{lccccc}\toprule
 & full labels & $90\%$ coarse & \multicolumn{2}{c}{novel classes (\%)} & fine classes\\
method (CUB official split) & acc. & acc. & superclass & leaf & exact leaf (\%)\\\midrule
Free-grained H-ViT + text attributes~\cite{park2026freegrain} (official code) & -- & .488 & 24.9 & 62.6 & --\\
Flat softmax + HSC & \textbf{.891} & .614 & 20.1 & 72.2 & 77.0\\
Conditional softmax & .888 & \textbf{.637} & 42.5 & 51.2 & 76.6\\
Per-level + fallback & .890 & .634 & 20.1 & 71.5 & 75.6\\
Uniform target & \textbf{.891} & .590 & 81.7 & 10.0 & 75.6\\
\rowcolor{ourspastel}\textbf{Arbor (ours)} & .885 & .630 & \textbf{84.6} & \textbf{9.1} & \textbf{81.8}\\
\bottomrule\end{tabular}
}
\caption{\textbf{CUB official split.} Free-grained recognition from its released code against the DINOv2 baselines and Arbor, on the same split, tree and protocols; three seeds. Novel classes are scored at the same operating point with each method's fallback rule. Exact leaf: exact-class rate on known-class images at the operating point, in the protocol with novel classes (not available for free-grained, which was also not run with full labels). Bold: column best.}\label{tab:offsplit_cmp}\end{table}
\section{Special cases: full tables}\label{app:full}
\Cref{tab:main} gives the special cases, \cref{tab:paired} the paired seed differences underlying their accuracy comparisons (the main protocol's are in \cref{tab:op}), and \cref{tab:special_full} every baseline with standard deviations.

\begin{table*}[!t]\centering\footnotesize\setlength{\tabcolsep}{3pt}
\resizebox{0.84\textwidth}{!}{\begin{tabular}{lccccccccccccccc}
\toprule
 & \multicolumn{3}{c}{Cars} & \multicolumn{3}{c}{CUB} & \multicolumn{3}{c}{CIFAR-100} & \multicolumn{3}{c}{Aircraft} & \multicolumn{3}{c}{iNat-2019} \\
\cmidrule(lr){2-4}\cmidrule(lr){5-7}\cmidrule(lr){8-10}\cmidrule(lr){11-13}\cmidrule(lr){14-16}
method & full & coarse & novel & full & coarse & novel & full & coarse & novel & full & coarse & novel & full & coarse & novel \\
\midrule
Flat softmax + HSC~\cite{goren2024hsc} & \textbf{.910} & .785 & 4 & .903 & .689 & 22 & .914 & .888 & 28 & \textbf{.836} & .696 & 5 & \textbf{.781} & .632 & 38 \\
Hierarchical CE~\cite{bertinetto2020making} & .910 & .736 & 21 & .902 & .693 & 40 & .914 & .859 & 51 & .830 & .692 & 21 & -- & -- & -- \\
Conditional softmax~\cite{redmon2017yolo9000,valmadre2022hierarchical} & .893 & .754 & 20 & .901 & .704 & 45 & .912 & .858 & 81 & .823 & .716 & 9 & .776 & .614 & 57 \\
Per-level + fallback~\cite{park2025hcast,garg2022haf,chu2026hieruq} & \textbf{.910} & .783 & 9 & \textbf{.904} & \textbf{.727} & 24 & .913 & \textbf{.893} & 33 & .832 & .694 & 9 & .779 & \textbf{.637} & 39 \\
HypStructure~\cite{zeng2024hypstructure} & .907 & .775 & 5 & .903 & .679 & 22 & \textbf{.914} & .876 & 11 & .829 & .687 & 7 & -- & -- & -- \\
\midrule
Uniform target & \textbf{.910} & .721 & 73 & .903 & .620 & 86 & .914 & .723 & 87 & \textbf{.836} & .713 & 35 & \textbf{.781} & .451 & 82 \\
\midrule
\rowcolor{ourspastel} \textbf{Arbor (ours)} & .902 & \textbf{.809} & \textbf{85} & .897 & .710 & \textbf{86} & .912 & .885 & \textbf{88} & .829 & \textbf{.718} & \textbf{78} & .746 & .593 & \textbf{86} \\
\bottomrule
\end{tabular}
}
\caption{\textbf{The special cases.} Full and coarse: known-class accuracy with all labels fine-grained, and with $90\%$ coarse labels and every class known. Novel: novel-class accuracy (\%) when every label is fine-grained except one novel class per superclass. Bold: column best among these methods, by unrounded value. On iNaturalist-2019, hierarchical CE and HypStructure were not run (--). Standard deviations and further baselines for the four small benchmarks in \cref{tab:special_full}.}
\label{tab:main}\end{table*}
\begin{table*}[!t]\centering\scriptsize\setlength{\tabcolsep}{4pt}\resizebox{0.62\textwidth}{!}{\begin{tabular}{lccc}\toprule
 & vs per-level + fallback & vs flat softmax over all nodes & vs flat softmax + HSC\\\midrule
Cars & $+2.6\pm0.4$ (min $+2.1$) & $+5.9\pm0.4$ (min $+5.4$) & $+2.4\pm0.4$ (min $+1.9$)\\
CUB & $-1.8\pm0.9$ (min $-2.7$) & $+5.2\pm1.4$ (min $+3.2$) & $+2.1\pm1.2$ (min $+0.5$)\\
CIFAR-100 & $-0.8\pm0.4$ (min $-1.4$) & $0.0\pm0.5$ (min $-0.4$) & $-0.3\pm0.2$ (min $-0.5$)\\
Aircraft & $+2.4\pm0.9$ (min $+1.5$) & $+0.6\pm0.3$ (min $+0.2$) & $+2.2\pm0.8$ (min $+1.2$)\\
\bottomrule\end{tabular}
}
\caption{\textbf{Paired seed differences in fine-grained accuracy at $q=0.9$ with every class known} (\cref{tab:main}; points). Each column is Arbor's validation-selected configuration minus a baseline on the same seed. The baselines are the per-level classifiers, the flat softmax over all nodes with its coarse-label weighting chosen on validation (\cref{tab:special_full}), and the flat softmax. Mean $\pm$ std and the minimum over the three seeds.}\label{tab:paired}\end{table*}
\paragraph{Results.} At $90\%$ coarse labels with every class known, Arbor is within $1.8$ points of the best single-model baseline on the four small benchmarks (above on Cars and Aircraft) and $0.5$--$1.8$ points behind the best five-member ensemble; with all labels fine-grained it is within $1$ point of the best single model; on iNat-2019 it is below by $3.5$--$4.4$ in both cases (\cref{tab:main}). With only the novel classes coarse, Arbor recognises $78$--$88\%$ of novel-class images; on the four small benchmarks the classification baselines of \cref{tab:merged} recognise $4$--$51\%$, except per-node sigmoids ($89\%$) and the conditional softmax ($81\%$) on CIFAR-100, and a flat softmax with a min/max weighting of its coarse labels recognises $91$--$94\%$ (\cref{tab:special_full}).
\begin{table*}[!t]\centering\scriptsize\setlength{\tabcolsep}{3pt}\resizebox{\textwidth}{!}{\begin{tabular}{lcccccccc}
\toprule
 & \multicolumn{4}{c}{Cars} & \multicolumn{4}{c}{CUB} \\
\cmidrule(lr){2-5}\cmidrule(lr){6-9}
method & $q{=}0$ & $q{=}0.9$ & novel & leaf & $q{=}0$ & $q{=}0.9$ & novel & leaf \\
\midrule
Flat softmax + HSC~\cite{goren2024hsc} & .910$\pm$0.002 & .785$\pm$0.004 & 4$\pm$1 & 87$\pm$1 & .903$\pm$0.006 & .689$\pm$0.012 & 22$\pm$1 & 71$\pm$0 \\
Deep ensemble ($5\times$) + HSC & .921$\pm$0.001 & .814$\pm$0.003 & 13$\pm$1 & 77$\pm$1 & .909$\pm$0.008 & .727$\pm$0.005 & 30$\pm$3 & 63$\pm$3 \\
Hierarchical CE~\cite{bertinetto2020making} & .910$\pm$0.000 & .736$\pm$0.005 & 21$\pm$1 & 72$\pm$1 & .902$\pm$0.009 & .693$\pm$0.019 & 40$\pm$0 & 56$\pm$1 \\
Soft labels~\cite{bertinetto2020making} & .910$\pm$0.002 & .786$\pm$0.009 & 8$\pm$1 & 84$\pm$1 & .900$\pm$0.008 & .693$\pm$0.013 & 20$\pm$1 & 77$\pm$1 \\
Conditional softmax~\cite{redmon2017yolo9000,valmadre2022hierarchical} & .893$\pm$0.002 & .754$\pm$0.009 & 20$\pm$1 & 69$\pm$1 & .901$\pm$0.007 & .704$\pm$0.008 & 45$\pm$1 & 49$\pm$2 \\
Per-node sigmoids~\cite{valmadre2022hierarchical} & .898$\pm$0.000 & .752$\pm$0.006 & 24$\pm$2 & 69$\pm$2 & .898$\pm$0.013 & .712$\pm$0.014 & 48$\pm$0 & 50$\pm$1 \\
Hyperbolic Busemann~\cite{atigh2021busemann} & .881$\pm$0.004 & .608$\pm$0.059 & 13$\pm$1 & 74$\pm$2 & .880$\pm$0.008 & .623$\pm$0.014 & 26$\pm$1 & 71$\pm$1 \\
Per-level + fallback~\cite{park2025hcast,garg2022haf,chu2026hieruq} & .910$\pm$0.003 & .783$\pm$0.005 & 9$\pm$0 & 77$\pm$1 & .904$\pm$0.008 & .727$\pm$0.010 & 24$\pm$2 & 67$\pm$3 \\
HypStructure~\cite{zeng2024hypstructure} & .907$\pm$0.001 & .775$\pm$0.008 & 5$\pm$1 & 85$\pm$0 & .903$\pm$0.007 & .679$\pm$0.009 & 22$\pm$3 & 71$\pm$3 \\
Uniform target & .910$\pm$0.002 & .721$\pm$0.012 & 73$\pm$3 & 16$\pm$3 & .903$\pm$0.006 & .620$\pm$0.013 & 86$\pm$1 & 6$\pm$1 \\
Deep ensemble, uniform target & .921$\pm$0.001 & .769$\pm$0.007 & 78$\pm$1 & 10$\pm$1 & .909$\pm$0.008 & .656$\pm$0.016 & 87$\pm$1 & 5$\pm$0 \\
Flat softmax + min/max weighting (variant and learning rate on validation) & .905$\pm$0.001 & .750$\pm$0.003 & 94$\pm$0 & 2$\pm$0 & .906$\pm$0.006 & .658$\pm$0.016 & 94$\pm$1 & 3$\pm$1 \\
Deep ensemble + min/max weighting, uniform target in superclasses without a novel class & .922$\pm$0.003 & .768$\pm$0.003 & 95$\pm$1 & 2$\pm$1 & .910$\pm$0.006 & .654$\pm$0.014 & 96$\pm$2 & 2$\pm$1 \\
\rowcolor{ourspastel} \textbf{Arbor (ours)} & .902$\pm$0.003 & .809$\pm$0.002 & 85$\pm$0 & 7$\pm$0 & .897$\pm$0.012 & .710$\pm$0.002 & 86$\pm$2 & 7$\pm$1 \\
\bottomrule
\end{tabular}
}\\[2mm]
\resizebox{\textwidth}{!}{\begin{tabular}{lcccccccc}
\toprule
 & \multicolumn{4}{c}{CIFAR-100} & \multicolumn{4}{c}{Aircraft} \\
\cmidrule(lr){2-5}\cmidrule(lr){6-9}
method & $q{=}0$ & $q{=}0.9$ & novel & leaf & $q{=}0$ & $q{=}0.9$ & novel & leaf \\
\midrule
Flat softmax + HSC~\cite{goren2024hsc} & .914$\pm$0.004 & .888$\pm$0.003 & 28$\pm$4 & 60$\pm$3 & .836$\pm$0.003 & .696$\pm$0.019 & 5$\pm$1 & 86$\pm$3 \\
Deep ensemble ($5\times$) + HSC & .919$\pm$0.003 & .893$\pm$0.004 & 60$\pm$4 & 29$\pm$5 & .848$\pm$0.003 & .693$\pm$0.018 & 5$\pm$1 & 86$\pm$2 \\
Hierarchical CE~\cite{bertinetto2020making} & .914$\pm$0.003 & .859$\pm$0.007 & 51$\pm$2 & 40$\pm$2 & .830$\pm$0.004 & .692$\pm$0.010 & 21$\pm$1 & 72$\pm$3 \\
Soft labels~\cite{bertinetto2020making} & .916$\pm$0.002 & .877$\pm$0.000 & 33$\pm$4 & 60$\pm$4 & .836$\pm$0.003 & .684$\pm$0.009 & 8$\pm$1 & 83$\pm$3 \\
Conditional softmax~\cite{redmon2017yolo9000,valmadre2022hierarchical} & .912$\pm$0.002 & .858$\pm$0.001 & 81$\pm$1 & 8$\pm$1 & .823$\pm$0.004 & .716$\pm$0.012 & 9$\pm$1 & 80$\pm$2 \\
Per-node sigmoids~\cite{valmadre2022hierarchical} & .911$\pm$0.003 & .875$\pm$0.002 & 89$\pm$1 & 5$\pm$1 & .822$\pm$0.006 & .668$\pm$0.009 & 25$\pm$1 & 68$\pm$3 \\
Hyperbolic Busemann~\cite{atigh2021busemann} & .907$\pm$0.001 & .846$\pm$0.010 & 29$\pm$1 & 65$\pm$2 & .835$\pm$0.003 & .675$\pm$0.009 & 13$\pm$2 & 76$\pm$3 \\
Per-level + fallback~\cite{park2025hcast,garg2022haf,chu2026hieruq} & .913$\pm$0.002 & .893$\pm$0.004 & 33$\pm$1 & 56$\pm$1 & .832$\pm$0.007 & .694$\pm$0.020 & 9$\pm$2 & 77$\pm$2 \\
HypStructure~\cite{zeng2024hypstructure} & .914$\pm$0.003 & .876$\pm$0.002 & 11$\pm$1 & 77$\pm$1 & .829$\pm$0.003 & .687$\pm$0.009 & 7$\pm$2 & 83$\pm$4 \\
Uniform target & .914$\pm$0.004 & .723$\pm$0.007 & 87$\pm$1 & 3$\pm$0 & .836$\pm$0.003 & .713$\pm$0.010 & 35$\pm$2 & 54$\pm$2 \\
Deep ensemble, uniform target & .919$\pm$0.003 & .833$\pm$0.003 & 87$\pm$1 & 3$\pm$0 & .848$\pm$0.003 & .716$\pm$0.006 & 37$\pm$4 & 52$\pm$3 \\
Flat softmax + min/max weighting (variant and learning rate on validation) & .913$\pm$0.001 & .886$\pm$0.001 & 94$\pm$1 & 2$\pm$0 & .832$\pm$0.008 & .712$\pm$0.010 & 91$\pm$2 & 3$\pm$1 \\
Deep ensemble + min/max weighting, uniform target in superclasses without a novel class & .919$\pm$0.002 & .831$\pm$0.002 & 94$\pm$1 & 2$\pm$0 & .846$\pm$0.005 & .735$\pm$0.012 & 92$\pm$2 & 2$\pm$1 \\
\rowcolor{ourspastel} \textbf{Arbor (ours)} & .912$\pm$0.003 & .885$\pm$0.005 & 88$\pm$1 & 5$\pm$1 & .829$\pm$0.006 & .718$\pm$0.010 & 78$\pm$3 & 13$\pm$2 \\
\bottomrule
\end{tabular}
}
\caption{\textbf{The special cases, every baseline, with standard deviations over three seeds.} $q{=}0$ and $q{=}0.9$: fine-grained accuracy with all labels fine-grained, and with $90\%$ coarse labels and every class known. Novel and leaf: with only the novel classes coarse, novel-class images assigned to their superclass or to a leaf (\%). Min/max weighting: one weight per superclass for the coarse labels, $\min_\ell \omega_\ell/\max_\ell \omega_\ell$ with $\omega_\ell=n_\ell/(n_\ell+\kappa)$, $\kappa$ a smoothing constant, applied to a flat softmax over all nodes.}\label{tab:special_full}\end{table*}
\end{document}